\documentclass[journal]{IEEEtran}
\usepackage{amsmath,amssymb}
\usepackage{algorithmic}
\usepackage{algorithm}
\usepackage{bbm}
\usepackage{tikz-cd}
\usepackage{array}
\usepackage[caption=false,font=normalsize,labelfont=sf,textfont=sf]{subfig}
\usepackage{textcomp}
\usepackage{stfloats}
\usepackage{url}
\usepackage{verbatim}
\usepackage{graphicx}
\usepackage{booktabs,color}
\usepackage{cite}
\usepackage[affil-it]{authblk}
\usepackage{stackengine} 
\usepackage{mathrsfs}
\usepackage{relsize}

\usepackage{amsmath}

\usepackage{indentfirst} 
\usepackage{multirow}
\usepackage{amssymb}
\usepackage{xcolor,eucal}
\usepackage{setspace}
\usepackage{tabularx}
\usepackage[colorlinks,linkcolor=red,anchorcolor=blue,citecolor=blue,CJKbookmarks=True]{hyperref}
\hypersetup{
    colorlinks=true,
    urlcolor=blue
}

\graphicspath{{figs/}}



\newcommand{\hanxiargmin}[1]{\mathop{\arg\min}\limits_{#1}}
\newcommand{\hanxiR}{\mathbb{R}}

\begin{document}
\title{Industrial Anomaly Detection and Localization Using Weakly-Supervised Residual
Transformers}
  
\author[1,2,$\dagger$]{Hanxi Li\thanks{\footnotesize This work was conducted during Hanxi Li's visit to Zhejiang University.}}
\author[1,2,$\dagger$]{Jingqi Wu}
\author[3]{Deyin Liu}
\author[4, $\star$]{Lin Wu}
\author[2]{Hao Chen}
\author[1]{Mingwen Wang}
\author[2, $\star$]{Chunhua Shen}

\affil[1]{Jiangxi Normal University, Jiangxi, China}
\affil[2]{Zhejiang University, Zhejiang, China}
\affil[3]{Anhui University, China}
\affil[4]{Swansea University, United Kingdom}
\affil[$\dagger$]{These authors contributed equally to this work}
\affil[$\star$]{Corresponding authors}

\markboth{Journal of \LaTeX\ Class Files,~Vol.~14, No.~8, February~2024}%
{Shell \MakeLowercase{\textit{et al.}}: A Sample Article Using IEEEtran.cls for IEEE Journals}

\maketitle
\begin{abstract} 


Recent advancements in industrial anomaly detection (AD) have demonstrated that incorporating a small number of anomalous samples during training can significantly enhance accuracy. However, this improvement often comes at the cost of extensive annotation efforts, which are impractical for many real-world applications. In this paper, we introduce a novel framework, ``\emph{Weak}ly-supervised \emph{RES}idual \emph{T}ransformer''
  (\emph{WeakREST}), designed to achieve high anomaly detection accuracy while minimizing the reliance on manual annotations. First, we reformulate the pixel-wise anomaly localization task into a block-wise classification problem. 
  Second, we introduce a residual-based feature representation called ``\emph{Pos}itional \emph{F}ast \emph{A}nomaly \emph{R}esiduals'' (\emph{PosFAR}) which captures anomalous patterns more effectively. To leverage this feature, we adapt the Swin Transformer for enhanced anomaly detection and localization. Additionally, we propose a weak annotation approach, utilizing bounding boxes and image tags to define anomalous regions. This approach establishes a semi-supervised learning context that reduces the dependency on precise pixel-level labels. To further improve the learning process, we develop a novel ResMixMatch algorithm, capable of handling the interplay between weak labels and residual-based representations.

On the benchmark dataset MVTec-AD, our method achieves an Average Precision (AP) of $83.0\%$, surpassing the previous best result of $82.7\%$ in the unsupervised setting. In the supervised AD setting, WeakREST attains an AP of $87.6\%$, outperforming the previous best of $86.0\%$. Notably, even when using weaker annotations such as bounding boxes, WeakREST exceeds the performance of leading methods relying on pixel-wise supervision, achieving an AP of $87.1\%$ compared to the prior best of $86.0\%$ on MVTec-AD. This superior performance is consistently replicated across other well-established AD datasets, including MVTec 3D and KSDD2. Code is available at: \url{https://github.com/BeJane/Semi_REST}

\end{abstract}


\section{Introduction} 
\label{sec:intro} 
\IEEEPARstart{P}{roduct} quality control is a critical aspect of modern manufacturing processes, and as a result, automatic defect inspection has become a highly sought-after solution in the manufacturing industry \cite{tao2022unsupervised, cao2023collaborative,10286557}. With sufficiently labeled training data, defect detection can be effectively performed using state-of-the-art image segmentation algorithms \cite{wang2022internimage, wang2022image}. However, real-world manufacturing scenarios often present a significant challenge: anomalous samples are substantially fewer than normal ones. This imbalance makes traditional supervised approaches less practical. To overcome this limitation, industrial defect detection is increasingly framed as an Anomaly Detection (AD) problem \cite{bergmann2019mvtec, mishra2021vt}, where only normal samples are used during training. This approach leverages the assumption that anomalies deviate from the learned representation of normality, enabling effective defect detection without relying on extensive labeled datasets of defective samples.
 
 \begin{figure}[t!]
 	\centering
\includegraphics[width=0.5\textwidth]{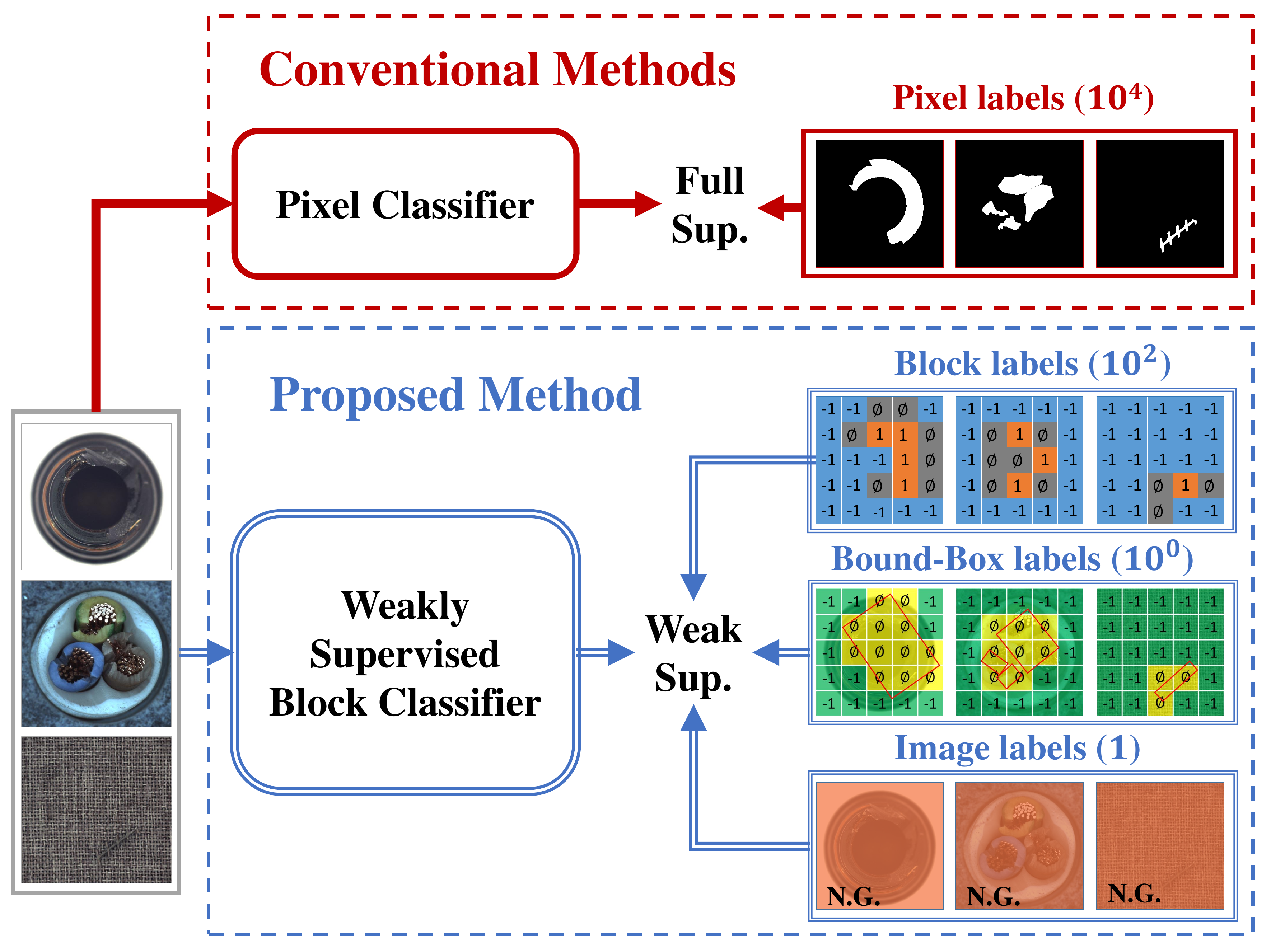}
 	\caption{
The comparison between the proposed weak annotation strategy and the conventional paradigm. Unlike traditional pixel-wise labels (see the top red box), our proposed annotations are categorized into three levels. Row-$1$: The AD problem is reformulated as block-wise binary classification. Normal samples, anomalous samples, and ignored samples are represented in blue, orange, and gray, respectively. This approach significantly reduces annotation granularity.
Row-$2$: A weaker labeling strategy using bounding boxes that encompass entire anomalous regions. This eliminates the need for pixel-level detail while still preserving key information about the defect.
Row-$3$: The weakest label using only tags indicating the defective status of the image. The numbers in the parenthesis denote the order of magnitudes (from $10^{4}$ to $1$) of the annotation clicks under the three levels of weak annotations. Best viewed in color.}
 \vspace{-0.2cm}
\label{fig:high_level}
 \end{figure}

The simplest approach to implementing anomaly detection involves classifying normal image patches as a single category and treating anomalous patches as outliers \cite{defard2021padim, zhang2022pedenet, chen2022deep}. To enhance anomaly localization, some researchers compare test image patches with normal references, either directly sourced from the training set \cite{roth2022towards, kim2022fapm, bae2022image, saiku2022enhancing, zhu2022anomaly, xie2023pushing} or reconstructed based on training samples \cite{shi2021unsupervised, hou2021divide, wu2021learning}. Furthermore, advanced techniques such as distillation-based methods \cite{9940966, deng2022anomaly, zhang2023destseg} and latent image registration frameworks \cite{huang2022registration, liu2023diversity} have been proposed to improve the precision and robustness of anomaly detection. These approaches leverage diverse strategies to capture subtle deviations from normal patterns, making them effective tools for industrial defect inspection.

Despite the effectiveness of the prevailing ``unsupervised'' learning methods, more
recent approaches \cite{ding2022catching, zhang2022prototypical,
yao2022explicit} show that introducing a small number of anomalous samples to the training process can lead to considerable performance gain. In practice,
obtaining a few abnormal samples for this ``supervised'' setting is feasible on a
continuously running assembly line. However, in this scenario, manually labeling pixel-wise anomalies is much more challenging which requires the annotator to instantly label each
new image to maintain the productivity of the assembly line.

In this paper, we present a novel anomaly detection (AD) framework that achieves a balance between high detection accuracy and reduced annotation cost. The high-level concept of the proposed approach is depicted in Fig.\ref{fig:high_level}. Specifically, we approach the AD task as a block-wise classification problem, significantly lowering the annotation burden by requiring only hundreds of labeled anomaly blocks on defective images instead of thousands of pixel-level annotations. At the core of our framework is a novel and efficient residual generation algorithm, termed ``Positional Fast Anomaly Residuals" (PosFAR), which generates robust features for each image block. These features are then classified as anomalous or normal using a Swin Transformer \cite{liu2021swin}. To further reduce annotation costs, we propose labeling anomaly regions with coarse labels such as bounding boxes or even image tags. As illustrated in rows 3 and 4  of Fig.~\ref{fig:high_level}, bounding boxes effectively enclose all anomaly regions, with the external blocks (in green) serving as normal samples for training. Meanwhile, internal blocks (in yellow), labeled as ``unknown" are leveraged using a novel semi-supervised learning algorithm specifically designed for our residual-based anomaly detector. This innovative use of unlabeled information enables our method to maintain high AD performance while relying on more economical annotations. The proposed algorithm, named ``\textbf{Weak}ly-supervised \textbf{RES}idual \textbf{T}ransformer'', i.e., ``WeakREST'', is validated on three benchmark datasets—MVTec-AD \cite{bergmann2019mvtec}, MVTec 3D \cite{bergmann2021mvtec}, and KSDD2 \cite{bovzivc2021mixed-KolektorSDD2} under both ``unsupervised" and ``supervised" settings. Experimental results demonstrate its superiority over state-of-the-art methods. Notably, WeakREST achieves superior performance even with only bounding-box annotations, outperforming existing methods that rely on stronger pixel-level supervisions.

The main contributions of this paper are as follows:
 \begin{itemize}
\item {\bf Annotation Efficiency:} Unlike conventional ``one-class" anomaly detection (AD) approaches, practical AD algorithms require more efficient annotation strategies. To address this, we introduce a novel annotation toolkit comprising block-wise labels, bounding box labels, and image-level labels. Notably, this is the first work to leverage block-wise labels for anomaly detection. Additionally, our innovative use of bounding box labels and image tags establishes a new paradigm for low-cost annotation in the AD literature.
   
\item {\bf Accuracy Advancement:} We propose the WeakREST algorithm, which incorporates a modified Swin Transformer \cite{liu2021swin} and a novel residual generation mechanism, namely ``PosFAR." Experimental results demonstrate that WeakREST consistently achieves superior performance, outperforming state-of-the-arts across three benchmark datasets under varying levels of supervision.

\item {\bf Enhanced Efficiency:} By utilizing inexpensive annotations such as bounding boxes and image-level tags, WeakREST effectively harnesses unlabeled features through the proposed ResMixMatch algorithm. Inspired by MixMatch \cite{berthelot2019mixmatch}, this approach designs a semi-supervised learning paradigm for the residual-based tokens. Remarkably, even with lightweight annotations, WeakREST surpasses SOTA methods that rely on costly pixel-level labels, thereby demonstrating both cost-effectiveness and efficiency.

\end{itemize}

The rest of this paper is organized as follows. Sec.~\ref{sec:related} presents the recent work related to this paper. The proposed method is detailed in Sec.~\ref{sec:method}.
Extensive experiments are conducted in Sec.~\ref{sec:experiments}, and
Sec.~\ref{sec:conclusion} concludes this paper.

\vspace{-0.1cm}
\section{Related Work}
\label{sec:related}

\subsection{ Industrial Anomaly Detection}
\label{subsec:supervision}
  
In the conventional setting of industrial anomaly detection tasks, all the training
    samples are anomaly-free and the defective patterns are detected as outliers in the
    test phase \cite{10251020,
    defard2021padim,zhang2022pedenet, chen2022deep,     
      lei2023pyramidflow}. This setting is usually referred
    to as ``unsupervised'' in the AD literature even though mild supervision, \emph{i.e.}
    the anomaly-free labels, still exist in the training set. To learn a discriminative
    model in this supervision condition, some sophisticated algorithms propose to generate
    artificial anomalous samples with synthetic defective regions \cite{
    zavrtanik2021draem, yang2023memseg, zhang2023destseg} for higher AD accuracy.  

Encouraged by the success of the AD models based on synthetic defects, a few methods \cite{ding2022catching, zhang2022prototypical, yao2022explicit} involved limited genuine anomalous samples to further unleash the discriminative power. They term this new setting as ``supervised'' in contrast to the default ``unsupervised'' setting. Note that in this supervision condition, the original
    anomaly-free samples as well as the fake defects are also employed in training. In this paper, we propose to replace the original pixel-level annotations with weak labels to reduce the annotation cost. We term this supervision condition as ``weakly-supervised'' and design a novel algorithm for leveraging the weak labels to achieve superior performance than the existing algorithms within the fully supervised condition. 

\subsection{Patch-Matching-based Anomaly Detection}

As a simple and typical example of the patch-match\-ing-based AD methods, PatchCore \cite{roth2022towards} proposes the coreset-subsampling algorithm to build a ``memory bank'' of patch features, which are obtained via smoothing the neutral deep
  features pre-learned on ImageNet \cite{deng2009imagenet, russakovsky2015imagenet}.  The
  anomaly score is then calculated based on the Euclidean distance between the test patch
  feature and its nearest neighbor in the ``memory bank''. Despite the simplicity,
  PatchCore performs dramatically well on the MVTec-AD dataset \cite{bergmann2019mvtec}.

Following the PatchCore \cite{roth2022towards}, PAFM \cite{kim2022fapm} applied patch-wise adaptive coreset
  sampling to ensure the efficiency. \cite{bae2022image} introduced the position and
  neighborhood information to refine the patch-feature comparison. G\-ra\-ph\-cor\-e
  \cite{xie2023pushing} utilized graph representation to customize Pa\-t\-ch\-Co\-re for
  the few-shot setting. \cite{saiku2022enhancing} modified Pa\-t\-ch\-Cor\-e by
  compressing the memory bank via k-means clustering. \cite{zhu2022anomaly} combined
  PatchCore \cite{roth2022towards} and Defect GAN \cite{zhang2021defect} for better outcome. Those methods are falling short of leveraging the intermediate information generated by the
  patch-matching. In this work, we use the matching residuals as the input tokens of our
  transformer model. The individual and the mutual information of the residuals are
  effectively exploited and SOTA performances are obtained.
        
\subsection{Swin Transformer for Anomaly Detection}
\label{subsec:swintransformer}
Swin Transformer
  \cite{liu2021swin, Liu_2022_CVPR} is variant of Vision Transformer (ViT) \cite{dosovitskiyimage}, which proposed a hierarchical Transformer with a shifted
  windowing scheme to introduces visual priors into Transformer with reduced computation cost. Swin Transformer has been deployed in various computer vision tasks, such as semantic segmentation
  \cite{huang2021fapn, cao2023swin}, instance segmentation
  \cite{dong2021solq,CTVIS} and object detection \cite{xu2021end,dai2021dynamic,liang2022cbnet}. 
 
In the field of anomaly detection (AD), the Swin Transformer has been widely explored as a backbone network. For example, \cite{uzen2022swin} introduces a hybrid decoder structure that integrates convolutional layers with the Swin Transformer, while \cite{gao2022cas} refines the original shifted windowing mechanism of the Swin Transformer for surface defect detection. Despite these advancements and the demonstrated success of Swin Transformer models in various domains, Swin-Transformer-based AD algorithms have struggled to consistently outperform state-of-the-art (SOTA) methods on benchmark datasets such as \cite{bergmann2019mvtec, mishra2021vt, bovzivc2021mixed-KolektorSDD2}. In this paper, we address the challenges posed by the small training datasets commonly encountered in AD tasks by adapting the Swin Transformer. Through a series of innovative modifications, we enhance both its performance and computational efficiency, making it better suited for the unique requirements of the AD domain.

\subsection{MixMatch and Weak Labels Based on Bounding Boxes}
\label{subsec:mixmatch_related}
    
Semi-supervised Learning (SSL) is attractive since it saves massive labeling labor. Many efforts have been devoted to utilizing the information from the unlabeled
  data \cite{ berthelot2019mixmatch, berthelot2019remixmatch,
  sohn2020fixmatch, wang2023freematch, WU-TCSVT}, mainly focusing on the generation of high-quality pseudo labels. Inspired by the seminar work \cite{zhangmixup,yun2019cutmix} for data augmentation, MixMatch proposes a multiple-loss SSL method that relies on a smart fusion  process between labeled and unlabeled samples and thus enjoys high accuracy and simplicity. 

In semantic segmentation, bounding boxes are usually used as weak supervision to reduce labeling costs \cite{CTVIS,WU-TIP,DeepSemantic-TMM}. \cite{hsu2019weakly} exploited the tightness prior to the bounding boxes to generate the positive and negative bags for multiple instance learning (MIL). \cite{kervadec2020bounding} integrated the tightness prior and a global background emptiness constraint derived from bounding box annotations into a weak semantic segmentation of medical images. \cite{lee2021bbam} proposed a bounding box attribution map (BBAM) to produce pseudo-ground-truth for weakly supervised semantic and instance segmentation. 

In this work, within the block-wise classification fra\-me\-wor\-k, MixMatch is smartly tailored to exploit the information of unlabeled blocks which are brought by the weak supervision of bounding boxes. This combination of the novel semi-supervised learning scheme and the bounding box labels is remarkably effective according to the experiment results and also novel in the literature, to our best knowledge.

\section{The Proposed Method}
\label{sec:method}

\subsection{Method Overview}
\label{subsec:infer}
\begin{figure*}[ht]  
  \centering
\includegraphics [width=0.95\textwidth]{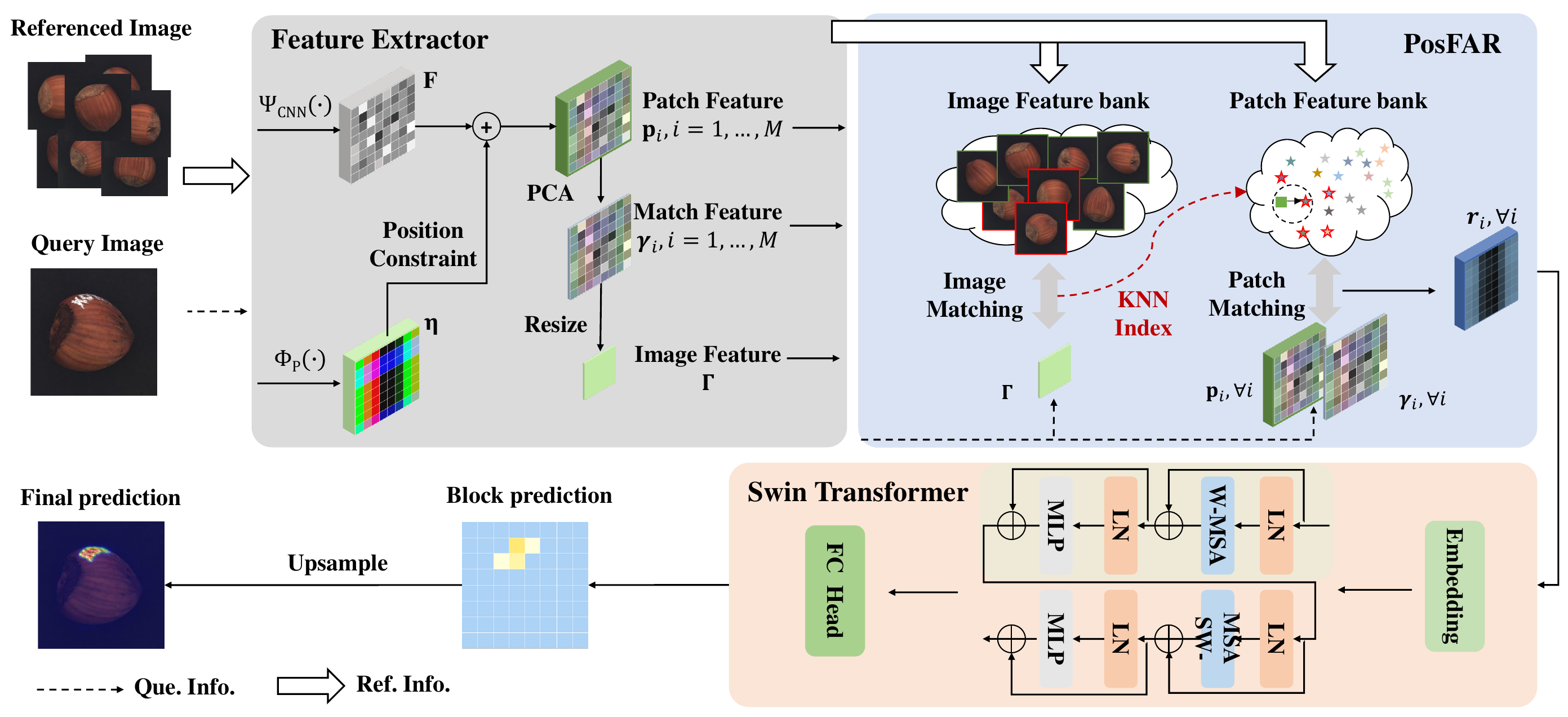}
\caption{ The overview of the INFERENCE process of WeakREST, which consists of three modules: feature extraction (see Sec.~\ref{subsec:PosFAR}),  PosFAR residual generator (see Sec.~\ref{subsec:PosFAR}) and Swin Transformer module for block-wise anomaly classification (see Sec.~\ref{subsec:swin}). In this
residual-based AD algorithm, the query information (from the test image) and reference information (from the training images) are utilized cooperatively to achieve high accuracy of anomaly detection and localization.
}
\vspace*{-0.4cm}
\label{fig:infer}
\end{figure*}

The overall inference process of our WeakREST algorithm is illustrated in
Fig~\ref{fig:infer}, and it considers the residual features of patch matching. The input contains the query (test) image and a set of reference images
which are defect-free. Three stages comprises the inference process: the
novel feature extracting stage and the residual feature (PosFAR) generation stage (the gray and blue boxes, see Sec.~\ref{subsec:PosFAR}); and the defect
classification process based on a Swin Transformer model (the orange box, see Sec.~\ref{subsec:swin}). 

\subsection{PosFAR: Fast Anomaly Residuals with P\-osi\-ti\-on Con\-s\-t\-r\-ai\-n\-ts}
\label{subsec:PosFAR}

\subsubsection{Matching Residual for Anomaly Detection}

Given an input image $\mathrm{I} \in \hanxiR^{h_{\mathrm{I}} \times
w_{\mathrm{I}} \times 3}$, one can extract deep features via  
\begin{equation}
  \label{equ:extract}
  [\mathbf{f}_1, \mathbf{f}_2, \cdots, \mathbf{f}_M] \gets \mathrm{Flatten} \gets \mathbf{F} =
  \Psi_{\mathrm{CNN}}(\mathrm{I}),
\end{equation}
where $\Psi_{\mathrm{CNN}}(\cdot)$ represents a deep neural network, which is pre-trained on a
large dataset (e.g., ImageNet \cite{russakovsky2015imagenet}).
$\mathbf{F} \in \hanxiR^{h_f \times w_f \times d_f}$ denotes the deep feature
tensor with $M$ feature vectors ($M = h_f\cdot w_f$). $\mathbf{f}_i \in \hanxiR^{d_f}, i =
1, \cdots, M$, stands for the $i$-th feature vector from the tensor $\mathbf{F}$. Then, we can build the memory ``bank'' of the defect-free training set via
\begin{equation}
  {\mathcal{B}}_{raw} = \{{\mathbf{f}}^{ref}_{i,j} \in \hanxiR^{d_f} \mid \forall j = 1, \cdots,
  N_{trn}, \forall i = 1, \cdots, M\},
\end{equation}
where $N_{trn}$ denotes the number of training images. ${\mathcal{B}}_{raw}$ contains
$M\cdot N_{trn}$ feature vectors, which is down-sampled as  
\begin{equation} 
  \label{equ:coreset}
  {\mathcal{B}} = \Psi_{\mathrm{core}}({\mathcal{B}}_{raw}) = \{{\mathbf{f}}^{ref}_t \in
  \hanxiR^{d_f} \mid \forall t = 1, \cdots, T\},
\end{equation}
where $\Psi_{\mathrm{core}}(\cdot)$ represents the ``coreset'' sampling scheme \cite{roth2022towards}, and $T \ll M\cdot N_{trn}$ bounds the matching complexity. Then, a test patch feature $\mathbf{f}^{tst}_i$ is matched against the reference
features in ${\mathcal{B}}$ via   
\begin{equation}
  \label{equ:pcr}
  t^{\ast} = \hanxiargmin{\forall t = 1, \cdots, T}\|\mathbf{f}^{tst}_i -
  {\mathbf{f}}^{ref}_t\|_{l_2}.
\end{equation}
The corresponding minimal distance $d_i = \|\mathbf{f}^{tst}_i -
{\mathbf{f}}^{ref}_{t^{\ast}}\|_{l_2}$ can be used to calculate the anomaly score of
the test patch \cite{roth2022towards, kim2022fapm, bae2022image, saiku2022enhancing}.
However, this vanilla version of patch matching suffers from information loss, low efficiency, and ignorance of patch locations. In
this paper, we introduce an effective patch-matching scheme for matching the residual
features, i.e., ``\textbf{Pos}itional \textbf Fast \textbf Anomaly \textbf Residuals'' (PosFAR). 

\subsubsection{Position Constrained Features}
\label{subsubsec:pcf}
As shown in \cite{bae2022image, gudovskiy2022cflow}, the positional information yielded by patch comparison
could improve AD performance. 
Herein, we are inspired by the ``positional embedding'' concept in the
Transformers \cite{dosovitskiyimage,liu2021swin} for patch matching: the original patch
features are aggregated with their positional features encoded in the Transformer way
\cite{he2022masked}. Given a patch feature $\mathbf{f}$ defined in
Eq.~\ref{equ:extract}, we generate its ``position-constrained '' version as  
\begin{equation}
  \label{equ:pcf}
  \mathbf{p} = \mathbf{f} + \lambda_{\mathrm{P}}\boldsymbol{\eta} =
  \mathbf{f} +\lambda_{\mathrm{P}}\Phi_{\mathrm{P}}(r, c),
\end{equation}
where $\boldsymbol{\eta} = \Phi_{\mathrm{P}}(r, c) \in \hanxiR^{d_f}$ is
termed Position Code \cite{he2022masked} whre the row-column coordinate
$[r, c]$ of $\mathbf{f}$ is extracted from the feature tensor $\mathbf{F} \in
\hanxiR^{h_f \times w_f \times d_f}$.  The function $\Phi_{\mathrm{P}}(\cdot)$ denotes the
positional embedding process that calculates the $k$-th element of $\boldsymbol{\eta}$ as 
\begin{equation}
  \label{equ:pe}
  \eta_k =
\begin{cases}
  \mathrm{sin}(\dfrac{c}{10000^{8k/d_f}})& k\in[0,\dfrac{d_f}{4})\\
  \mathrm{cos}(\dfrac{c}{10000^{8(k-d_f/4)/d_f}})& k\in[\dfrac{d_f}{4},\dfrac{d_f}{2})\\
  \mathrm{sin}(\dfrac{r}{10000^{8(k-d_f/2)/d_f}})& k\in[\dfrac{d_f}{2},\dfrac{3d_f}{4})\\
  \mathrm{cos}(\dfrac{r}{10000^{8(k-3d_f/4)/d_f}})& k\in[\dfrac{3d_f}{4},d_f),
\end{cases}
\end{equation}
where $k \in [1, d_f]$, $r \in [1, h_f]$, $c \in [1, w_f]$.
%
The resultant patch feature matching  is constrained by the positional
information and the new patch feature is termed ``Position Constrained Feature'' (PCF). 
The ablation study in Section~\ref{subsec:Ablation_study} verifies the merit of
this constraint. 

\subsubsection{Matching in a Low Dimensional Space}
\label{subsubsec:far}
Matching patch features in their original space is time-consuming due to high dimensionality. To this end, we propose to generate low-dimensional anomaly residuals with high discriminant. First, the patch matching is performed in a lower-dimensional space. In specific, a Principle Component Analysis (PCA) is conducted over the learned PCFs, and each feature $\mathbf{p} \in \hanxiR^{d_f}$ is mapped to
an lower-dimensional space as $\boldsymbol{\gamma} = \Psi_{\text{PCA}}(\mathbf{p}) \in
\hanxiR^{d_l}$, where $d_l \ll d_f$. 
In this way, one can convert the bank $\mathcal{B}$ defined in
Eq.~\ref{equ:coreset} into its position-constrained and lower-dimensional version as
$\mathcal{B}^P_l = \{{\boldsymbol{\gamma}}^{ref}_t \in \hanxiR^{d_l} \mid
\forall t = 1, \cdots, T\}$. In terms of the $i$-th ($i \in [1, 2, \cdots, h_f\cdot w_f]$)
test patch, the patch matching can be performed efficiently in its lower-dimensional space via 
\begin{equation}
  \label{equ:fast}
  t^{\ast} = \hanxiargmin{\forall t = 1, \cdots, T}\|\boldsymbol{\gamma}^{tst}_i -
  {\boldsymbol{\gamma}}^{ref}_t\|_{l_2},
\end{equation}
where $\boldsymbol{\gamma}^{tst}_i = \Psi_{\text{PCA}}(\mathbf{p}^{tst}_{i})$ represents
the lower-dimensional PCF of the test patch.

\subsubsection{Matching with Similar Reference Images}
\label{subsubsec:similar}
To further accelerate the matching process, we propose to \emph{match a test patch only with
the reference patches ``similar'' to reference images.} To quantify the image similarity, we first generate the image feature of an image $\mathrm{I}$
as
\begin{equation}
\begin{tikzcd}
  \mathbf{F} \in \hanxiR^{h_f \times w_f \times d_f}
  \arrow[r, "+\lambda_P\boldsymbol{\eta}"]
   & 
  \mathbf{P} \in \hanxiR^{h_f \times w_f \times d_f} 
  \arrow[d, "\Psi_{\text{PCA}}(\cdot)"] 
  \\ 
  \boldsymbol{\Gamma} \in \hanxiR^{h_l \times w_l \times d_l}
  & 
  \mathbf{P}_l \in \hanxiR^{h_f \times w_f \times d_l}, 
  \arrow[l, "\mathrm{resize}"]
\end{tikzcd}
\end{equation}
where $\mathbf{F} = \Psi_{\mathrm{CNN}}(\mathrm{I})$ is defined in Eq.~\ref{equ:extract},
$\mathbf{P}$ denotes the feature tensor containing $h_f\cdot w_f$ PCFs. The ``resize'' operation can reduce the width and height of the feature tensor via interpolation. Then the distance between the $j$-th reference image $\mathrm{I}^{ref}_j$ and the test image
$\mathrm{I}^{tst}$ is defined as
\begin{equation}
  \label{equ:robust_dist}
  \delta_j = \Delta(\mathrm{I}^{tst}, \mathrm{I}^{ref}_j), \forall j \in 1, \cdots, N_{trn},
\end{equation}
where $\Delta(\cdot)$ denotes the ``robust image distance'' \cite{li2023target}.
Given all the distances between $\mathrm{I}^{tst}$ and the reference images, referred as $\{\delta_1, \delta_2, \cdots, \delta_{N_{trn}}\}$, the image indexes of
$\mathrm{I}^{tst}$'s ``similar reference images'' are defined as
\begin{equation}\label{equ:knn}
 \resizebox{.88\hsize}{\height}{
$  \{\delta_1, \delta_2, \cdots, \delta_{N_{trn}}\} \xrightarrow{K\texttt{-NN Indexes}}
   \mathcal{Q} = \{q_1, q_2, \cdots, q_K\}.$
   }
\end{equation}


\subsubsection{Generating PosFAR}
Given a test image $\mathrm{I}^{tst}$ and a set of defect-free training images
$\{\mathrm{I}^{ref}_1, \mathrm{I}^{ref}_2, \cdots, \mathrm{I}^{ref}_{N_{trn}}\}$, we can generate the reference bank $\mathcal{B}^P = \{{\mathbf{p}}^{ref}_t \in
\hanxiR^{d_f} \mid \forall t = 1, \cdots, T\}$ and its low-dimensional correspondence
$\mathcal{B}^P_l = \{\boldsymbol{\gamma}^{ref}_t \in \hanxiR^{d_l} \mid \forall t = 1,
\cdots, T\}$. Meanwhile, the training image index of each element in $\mathcal{B}^P$ is saved in the set $\{j_1, j_2, \cdots, j_T\}$.
The proposed faster patch matching can be defined as 
\begin{equation}
  \label{equ:faster}
  t^{\ast} = \hanxiargmin{\forall j_t \in \mathcal{Q}}\|\boldsymbol{\gamma}^{tst}_i -
  {\boldsymbol{\gamma}}^{ref}_t\|_{l_2}, ~\forall i = 1, \cdots, M,
\end{equation}
where $\mathcal{Q}$ denotes the $K$-NN indexes of $\mathrm{I}^{tst}$ as defined in
Eq.~\ref{equ:knn}, $\boldsymbol{\gamma}^{tst}_i$ stands for the lower-dimensional PCF
feature of $i$-th patch in $\mathrm{I}^{tst}$. Finally, the PosFAR feature of each test patch is calculated via
\begin{equation}
\label{equ:posfar}
  \mathbf{r}_i = \lceil \text{ABS}(\mathbf{p}^{tst}_i - {\mathbf{p}}^{ref}_{t^{\ast}})
  \rceil^{\theta} \in \hanxiR^{d_f}, ~ \forall i,
\end{equation}
where $\text{ABS}(\cdot)$ denotes the function of absolute value, $\lceil \cdot
\rceil^{\theta}$ stands for the element-wise $\theta$-power operation which outweighs the higher values in the residual vector. Compared with the distance-based residuals \cite{roth2022towards, kim2022fapm, bae2022image, saiku2022enhancing}, PosFAR contains much richer information for patch matching. Each $\mathbf{r}_i$ represents an ``image block'' which can be easily recognized as
defective or defect-free by using the Swin transformer described below.
 \vspace*{-0.3cm}

\subsection{ Residual-based Swin Transformer for Block-wise Anomaly Detection}\label{subsec:swin}

\subsubsection{Block-wise Anomaly Labels}\label{sec:block_label}

Inspired by \cite{roth2022towards, zhang2022prototypical,
yang2023memseg, zhang2023destseg}, we employ a discriminative model to predict the anomaly score map for test images. In the conventional ``unsupervised''
setting (as described in Sec.~\ref{subsec:supervision}), pseudo defective regions are
usually generated so that the segmentation model \cite{roth2022towards,
zhang2022prototypical, yang2023memseg, zhang2023destseg} can be trained properly with the
pixel-wise labels. 
However, as the proposed PosFAR feature is block-wise, we
propose to cast the original pixel-wise segmentation task into a block classification problem. 
Accordingly, the pixel labels need to be converted into the block labels. 

\begin{figure}[!t]  
  \centering
\includegraphics [width=0.3\textwidth]{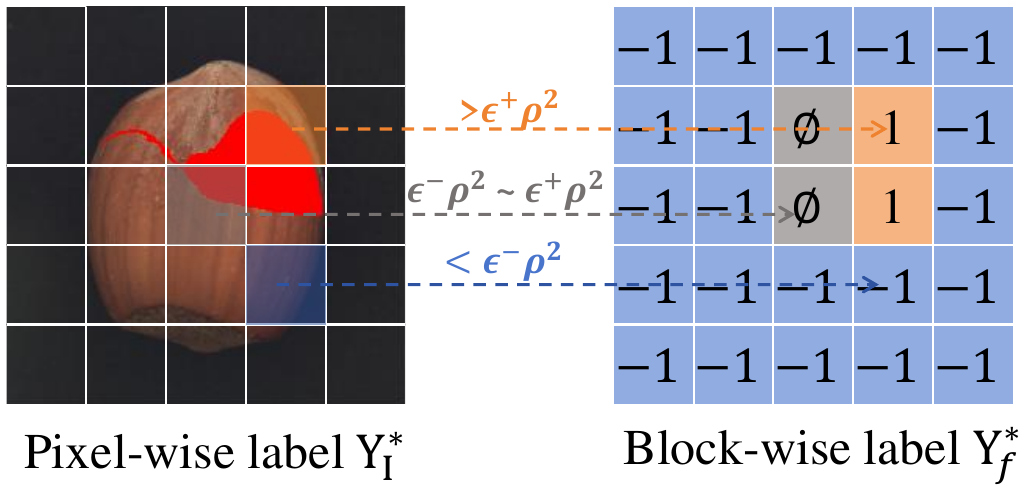}
\caption{
The block labeling strategy. The blocks with more than
  $\epsilon^{+}\rho^2$ anomaly pixels are labeled $1$ (red) while those blocks with less than
  $\epsilon^{-}\rho^2$ are labeled $-1$ (blue). The remaining blocks are labeled
  $\emptyset$ and will be ignored in the training phase.
}
\vspace*{-0.3cm}
\label{fig:block_label}
\end{figure}

Suppose that the pixel label map of an image $\mathrm{I} \in \hanxiR^{h_{\mathrm{I}}
\times w_{\mathrm{I}} \times 3}$ is denoted as $\mathrm{Y}^{\ast}_{\mathrm{I}} \in
\hanxiR^{h_{\mathrm{I}} \times w_{\mathrm{I}}}$ (see Fig.~\ref{fig:block_label}), with $0$ indicating defect-free pixels
while $1$ stands for the anomalous ones. 
We then can define our block-wise label map $\mathrm{Y}^{\ast}_{f} \in
\hanxiR^{h_f \times w_f}$ as
\begin{equation}
  \label{equ:block_label}
  ~ \mathrm{Y}^{\ast}_{f}(r_f, c_f) =
  \begin{cases}
    1 & \mathlarger{\mathlarger{\sum}}\limits_{(r_{\mathrm{I}}, c_{\mathrm{I}}) \in  \mathcal{U}_{r_f,
    c_f}}\mathrm{Y}^{\ast}_{\mathrm{I}}(r_{\mathrm{I}}, c_{\mathrm{I}}) > \epsilon^{+}\rho^2 \\
    -1 & \mathlarger{\mathlarger{\sum}}\limits_{(r_{\mathrm{I}}, c_{\mathrm{I}}) \in  \mathcal{U}_{r_f,
    c_f}}\mathrm{Y}^{\ast}_{\mathrm{I}}(r_{\mathrm{I}}, c_{\mathrm{I}}) < \epsilon^{-}\rho^2\\
    \emptyset & \text{otherwise}
  \end{cases}
\end{equation}
where $\mathcal{U}_{r_f, c_f}$ denotes the pixels belonging to the image block at
coordinate $[r_f, c_f]$; $\rho = h_{\mathrm{I}}/{h_f} = w_{\mathrm{I}}/{w_f}$;
$\epsilon^{+}$ and $\epsilon^{-}$ are the two predefined thresholds; when labeled as
$\emptyset$, the block is ignored during training, as introduced in
Sec.~\ref{subsubsec:swin}. Fig.~\ref{fig:block_label} illustrates this block labeling
scheme. Note that this process for the synthetic anomalies is conducted automatically 
and requires NO manual annotation. 
   
\emph{In this work, the block-wise labels are employed for the synthetic defects in the
``unsupervised'' setting as well as the genuine defects in the ``supervised'' setting.}
The experimental results of this work verify the superiority of this labeling strategy. On
the other hand, in real-life AD tasks, one only needs to manually label image blocks
rather than pixels and thus significant reduction on annotation cost is achieved. 

\subsubsection{Swin Transformer with Focal Loss}
\label{subsubsec:swin}
We convert the AD task into a block-wise binary classification problem and solve it by
using a Swin Transformer model \cite{liu2021swin}. In specific, given a test image
$\mathrm{I}^{tst} \in \hanxiR^{h_{\mathrm{I}} \times w_{\mathrm{I}} \times 3}$, its
PosFARs $\mathbf{r}_i \in \hanxiR^{d_f}, \forall i$ are calculated via
Eq.~\ref{equ:posfar} then fed into the Swin Transformer model as the input tokens
\cite{dosovitskiyimage, liu2021swin}. 

Given that anomaly detection is often performed on relatively small datasets \cite{bergmann2019mvtec, mishra2021vt}, we propose a compact and efficient miniature Swin Transformer model. As illustrated in Fig.~\ref{fig:infer}, the pipeline begins with a linear embedding layer applied to the PosFAR features, projecting them into a 1024-dimensional space. This is followed by four Swin Transformer blocks, which leverage a 32-head self-attention mechanism within $8\times 8$ regular windows (W-MSA) and shifted windows (SW-MSA). Finally, each token, representing an image block, is classified as either normal or anomalous using a fully connected layer. Mathematically we define 
\begin{equation}
  p^i_{+} = \Psi_{\mathrm{Swin}}(\mathbf{r}_i), ~ \forall i \in [1, 2, \cdots, M],
\end{equation}
where $p^i_+ \in [0, 1]$ denotes the normalized anomaly confidence (of the $i$-th token)
predicted by the Swin Transformer $\Psi_{\mathrm{Swin}}(\cdot)$

Considering that the normal image blocks usually dominate the original data distribution,
we employ the focal loss \cite{lin2017focal} to lift the importance of the anomaly class.
The focal loss used in this work writes:
\begin{equation}
  \label{equ:focal}
  \begin{split}
    \mathcal{L}_{F} = & -\dfrac{1}{|\mathcal{Z}^{-}|}\sum\limits_{i \in \mathcal{Z}^{-}}\left[ (1-\alpha)
    {p_{+}^i}^{\gamma}\mathrm{log}(1 - p^i_{+})\right] \\
    & -\dfrac{1}{|\mathcal{Z}^{+}|} \sum\limits_{i \in \mathcal{Z}^{+}}\left[ \alpha
    (1-p^i_{+})^{\gamma}\mathrm{log}(p^i_{+})\right]
  \end{split}
\end{equation} where $\mathcal{Z}^{+}$ and $\mathcal{Z}^{-}$ stands for the training
sample sets (here are transformer tokens) corresponding to defective ($+$) and defect-free
($-$) classes respectively. 

\subsubsection{Randomly Masked Residuals}
\label{subsubsec:mask}

Different from the vanilla vision transformers \cite{dosovitskiyimage, liu2021swin,
Liu_2022_CVPR}, the input to our model is essentially \textit{feature residual vectors}.
Most conventional data augmentation methods \cite{shorten2019survey, yun2019cutmix,
yang2022image} designed for images can not be directly used in the current situation.
By contrast, inspired by the
recently proposed MAE algorithm \cite{he2022masked}, we design a simple but effective feature
augmentation approach termed ``Randomly Masked Residuals'' for achieving higher generalization
capacity. In specific, when training, each tokens $\{\mathbf{r}_1, \mathbf{r}_2, \cdots,
\mathbf{r}_M\}$ defined in Eq.~\ref{equ:posfar} is randomly ``masked'' or ``noised'' as
\begin{equation}
  \label{equ:dropout}
  \forall i, ~\mathbf{r}_i = 
  \begin{cases}
    \mathbf{0}^{\mathrm{T}} \in \hanxiR^{d_f} ~ & \tau \in [0, \beta] \\
     \mathbf{r}_i + \kappa \frac{\|\mathbf{r}_i\|_{l_2}}{\|\mathbf{g}\|_{l_2}}\mathbf{g} ~
     & \tau \in (\beta, 1] \\
  \end{cases}
\end{equation}
where $\tau$ is a random variable sampled from the uniform distribution $[0, 1]$; $\beta$
is the constant controlling the frequency of the reset operation; $\mathbf{g} \in
\hanxiR^{d_f}$ denotes a Gaussian noise vector; $\kappa \in [0, 1]$ is a small constant
for residual jittering. 

\subsubsection{Off-the-shelf methods for generating fake anomalies} 
In the ``unsupervised'' setting of AD tasks, one needs to generate fake anomalies to train a
discriminative model properly. In this work, we follow the off-the-shelf fake/simulated
anomaly generation approach proposed in the MemSeg algorithm \cite{yang2023memseg}.
Readers are recommended to the original work \cite{yang2023memseg} for more details. Note
that we also employ this anomaly generation method for the supervised and
weakly-supervised settings to increase the variation of the training samples.

\subsubsection{Inference of Swin Transformer} 
Given $\{p^1_{+}, p^2_{+},$ $\cdots, p^M_{+}\}$ standing for the anomaly confidences of
image blocks predicted by the Swin Transformer model $\Psi_{\mathrm{Swin}}(\cdot)$, one
can obtain the image-size anomaly map $\mathbf{P}^{\ast}_{+} \in \hanxiR^{h_I \times w_I}$
as
\begin{equation}
  \label{equ:final_pred}
  \{p^1_{+}, \cdots, p^M_{+}\} \xrightarrow{\texttt{reshape}} \mathbf{P}_{+} \in \hanxiR^{h_f
\times w_f} \xrightarrow{\texttt{upsample}} \mathbf{P}^{\ast}_{+}
\end{equation}

\subsection{Exploiting the Unlabeled Information via ResMixMatch}
\label{subsec:mixmatch}

\subsubsection{ Weaker Labels with Minimal Labeling Cost}
\label{sec:weaker}
\begin{figure}[t]  
  \centering
\includegraphics [width=0.35\textwidth]{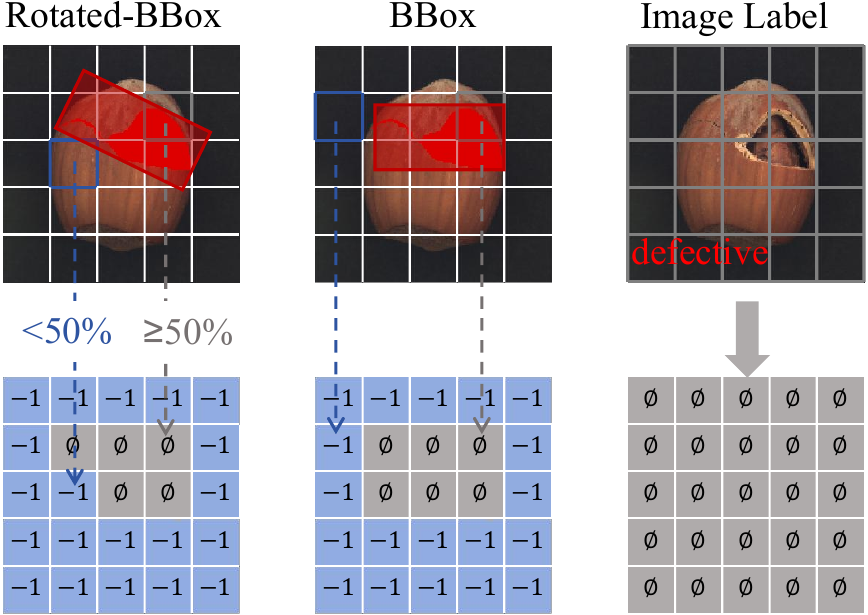}
\caption{Three types of weak labels considered in this paper. From left to right: the``rotated bounding-boxes'' (left), the ``axis-aligned bounding-boxes'' (middle) and the ``image-level labels'' (right). The lower part of each column illustrates the block-wise label conversion for the corresponding weak label.}
  \vspace*{-0.2cm}
\label{fig:weaker}
\end{figure}

To further reduce the annotation cost, we introduce three types of anomaly labels which need less labeling costs than the block-wise ones, e.g., \cite{Tabernik2019JIM-KolektorSDD, bovzivc2021mixed-KolektorSDD2} show that
bounding boxes can be deployed to annotate defective parts. As depicted in Fig.~\ref{fig:weaker}, we consider three weak labels: ``rotated bounding-boxes'' (left), ``axis-aligned bounding-boxes'' (middle) and ``image-level labels'' (right). Also, Fig. \ref{fig:weaker} shows that bounding box labels are the minimal rectangles covering the whole defective
region, with or without rotation. On the other hand, the image-level label just
represents the defective status of the image. These weak labels only requires the
annotators to supply a few ($1$ to $4$) clicks on the image. 

In the block-based development, one needs to convert the weak labels into corresponding block-wise annotations to suit the training. The lower part of Fig.~\ref{fig:weaker}
illustrates such converting processes for three levels of weak labels. In a
nutshell, for the bounding-box-based labels, we consider the outside blocks (overlapping
ratio greater than $50\%$) of the bounding boxes as normal while the inside blocks
(overlapping ratio less than $50\%$) are treated as ``unknown''. Nonetheless, all the blocks of an image labeled as defective are unknown.


\subsubsection{A Novel Semi-Supervised Learning Paradigm}

\begin{figure*}[ht]  
  \centering
\includegraphics [width=0.8\textwidth]{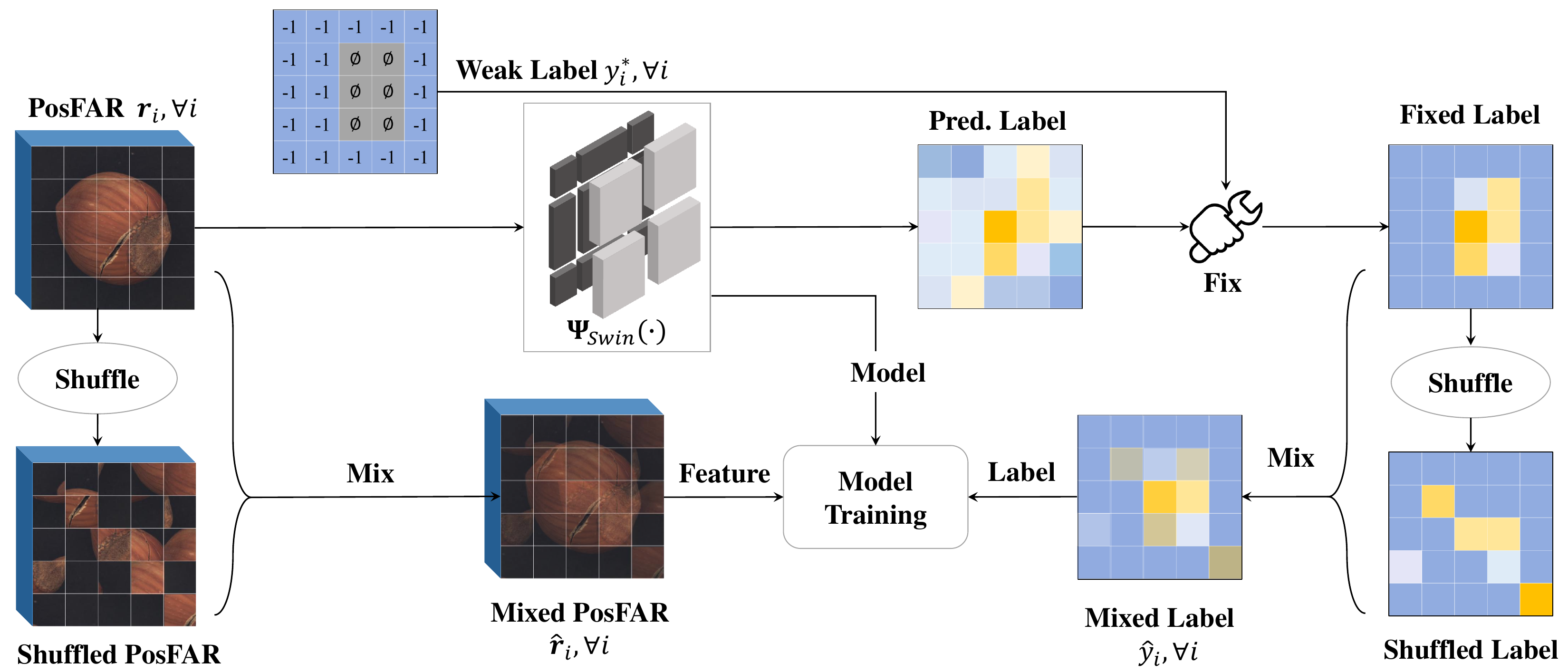}
\caption{
  The overview of the ResMixMatch training paradigm. The weak label only defines the non-defective
  region and the unknown region. It is used to ``fix'' the estimated label predicted by
  the Swin-Transformer model $\Psi_{swin}(\cdot)$. As the name suggests, the proposed
  ResMixMatch algorithm train its network model by using the ``mixed'' labels and
  residuals. 
}
\label{fig:resmixmatch}
\end{figure*}

The proposed weaker labels lead to very efficient annotation processes. However, they
arise another difficulty in training: An large portion of the image blocks are unlabeled.
Fortunately, this semi-supervised situation is well studied in the machine learning
literature \cite{berthelot2019mixmatch, berthelot2019remixmatch, sohn2020fixmatch,
wang2023freematch}. In this work, we introduce the high-level concept of the MixMatch
algorithm \cite{berthelot2019mixmatch} into the learning process of a our WeakREST model.
The yielded semi-supervised learning algorithm, termed ``ResMixMatch'', is specifically
designed for our residual-learning scenario. The workflow of ResMixMatch is depicted in
Fig.~\ref{fig:resmixmatch}. As we can see, the weak label $y^{\ast}_i, \forall i$ is used
to ``fix'' the estimated labels predicted by the Swin-Transformer. Similar to MixMatch
\cite{berthelot2019mixmatch}, our network model $\Psi_{swin}(\cdot)$ is trained by using
the ``mixed'' labels and residuals. On the other hand, different from MixMatch
that treats every sample independently, in ResMixMatch, all the PosFAR features are
related. The Swin Transformer model can effectively link the PosFARs from the same image
via the self-attention mechanism and their anomaly confidences are then predicted
depending on each other. The labels of the ``unknown'' blocks are estimated not only by
the mixing-matching strategy but also based on the neighboring information. In this way,
the ``label guessing'' becomes more confident. To formally illustrate ResMixMatch, we
summarize the proposed semi-supervised learning paradigm in Algorithm~\ref{alg:mixmatch}.

\begin{algorithm}[!ht]
\setstretch{0.48} 
\caption{ResMixMatch training of WeakREST}
\begin{algorithmic}[1]
\STATE \textbf{Input:}
  {
    Swin Transformer model $\Psi_{\mathrm{Swin}}(\cdot)$, PosFARs $\mathbf{r}_i \in \hanxiR^{d_f}$, the
    corresponding labels $y^{\ast}_i  \in \{\text{-}1,\emptyset(\text{unknown})\}, i = 1, 2, \cdots,
M$, 
    sharpening temperature $t$, unlabeled loss weight $\lambda_u$, number of augmentations $A$, and focal loss
    parameters $\{\alpha_x, \alpha_u, \gamma_x,\gamma_u \}$.
  }

\STATE $A$-Augmentation as Eq.~\ref{equ:dropout}  
   {\begin{align*}
  & \forall i,~  \{\mathbf{r}_{i,j}, \forall j \mid j = 1, 2, \cdots, A\} \gets A\text{-Augmentation} \gets \mathbf{r}_i\\
  & \forall i, ~ \{y^{\ast}_{i,j}, \forall j \mid j = 1, 2, \cdots, A\} \gets \text{Copy}
  \gets y^{\ast}_i,
 \end{align*}} 
\STATE Guess pseudo labels through augmentation 
 {\begin{align*}
   &  \forall i,~  \{\bar{y}_{i,j}, \forall j \mid j = 1, 2, \cdots, A \}\gets  \text{Copy} \\
 &  \gets \mathrm{Sharpen}\left(\dfrac{1}{A}  \sum^{A} \limits_{j=1}
  \Psi_{\mathrm{Swin}}(\mathbf{r}_{i,j}), t\right)
 \end{align*}}
\STATE   Divide the tokens into labeled set $\mathcal{X}$ and unlabeled set $\mathcal{U}$ 
 { \begin{align*}
 &  \mathcal{X} = \{X_i = \{\mathbf{r}_i, y^{\ast}_i\}, \forall i \mid y^{\ast}_i = \text{-}1\}\\
   & \mathcal{U} = \{U_i = \{\mathbf{r}_i, \bar{y}_i\},  \forall i \mid y^{\ast}_i = \emptyset\} 
  \end{align*}}
 \STATE Combine the labeled and unlabeled tokens and shuffle
  { \begin{align*}
    \mathcal{W} = \mathrm{Shuffle}(\mathrm{Union}(\mathcal{X},\mathcal{U})) 
   \end{align*}}
\STATE Apply MixUp \cite{berthelot2019mixmatch} to all tokens
  {  \begin{align*} 
 &  \hat{\mathcal{X}} \gets \{\mathrm{MixUp}(X_i, W_i), \forall i \mid i =
  1,\cdots,|\mathcal{X}|\} \\
  & \hat{\mathcal{U}} \gets \{\mathrm{MixUp}(U_i, W_{i+|\mathcal{X}|}), \forall i \mid i =
  1,\cdots,|\mathcal{U}|\} 
  \end{align*} }
\STATE Randomly mask tokens as Eq.~\ref{equ:dropout}
{ \begin{align*}
\forall \{ \hat{\mathbf{r}}_i,\hat{y}_i \} \in Union(\hat{\mathcal{X}},\hat{\mathcal{U}}),~ \{ \hat{\mathbf{r}}_i,\hat{y}_i \} =
  \mathrm{RandomMask}(\hat{\mathbf{r}}_i,\hat{y}_i)
    \end{align*}} 
\STATE   Classify the tokens 
{  \begin{align*}
  \forall i,~ {p}^i_{+} = \Psi_{\mathrm{Swin}}(\hat{\mathbf{r}}_i)
  \end{align*}} 
\STATE Compute the labeled loss $\mathcal{L}_x$ and unlabeled loss $\mathcal{L}_u$
{  
\begin{align*}
  & \mathcal{Z}^{+}_{\mathrm{k}} = \{\forall i \mid y_i^{\ast} =\text{-}1 ~ \& ~ \hat{y}_i >
  0.5 \}\\
  & \mathcal{Z}^{+}_{\mathrm{u}} = \{\forall i \mid y_i^{\ast} =\emptyset~\& ~
  \hat{y}_i > 0.5 \}\\
  & \mathcal{Z}^{-}_{\mathrm{k}} = \{\forall i \mid y_i^{\ast} =\text{-}1~\& ~\hat{y}_i
  \leqslant 0.5 \}\\
  &\mathcal{Z}^{-}_{\mathrm{u}} = \{\forall i \mid y_i^{\ast} =\emptyset~ \& ~
  \hat{y}_i \leqslant 0.5 \}\\
  &\mathcal{L}_{x} =  -\dfrac{1}{|\mathcal{Z}_k^{-}|}\sum\limits_{i \in \mathcal{Z}_k^{-}}\left[ (1-\alpha_x)
    {p_{+}^i}^{\gamma_x}\mathrm{log}(1 - p^i_{+})\right] \\
  &  -\dfrac{1}{|\mathcal{Z}_k^{+}|} \sum\limits_{i \in \mathcal{Z}_k^{+}}\left[ \alpha_x
    (1-p^i_{+})^{\gamma_x}\mathrm{log}(p^i_{+})\right]\\
  &\mathcal{L}_{u} =  -\dfrac{1}{|\mathcal{Z}_u^{-}|}\sum\limits_{i \in \mathcal{Z}_u^{-}}\left[ (1-\alpha_u)
    {p_{+}^i}^{\gamma_u}\mathrm{log}(1 - p^i_{+})\right] \\
  &  -\dfrac{1}{|\mathcal{Z}_u^{+}|} \sum\limits_{i \in \mathcal{Z}_u^{+}}\left[ \alpha_u
    (1-p^i_{+})^{\gamma_u}\mathrm{log}(p^i_{+})\right]    
\end{align*}
} 
\STATE \textbf{Output:} {$\mathcal{L}_{mix} = \mathcal{L}_x + \lambda_u \mathcal{L}_u $ }

  \vspace{-0.1cm}
  \end{algorithmic}

  \label{alg:mixmatch}
\end{algorithm}

\vspace{-0.1cm}
\subsection{Fast Foreground Region Estimation}
\label{subsec:fore}

Recent researches on industrial anomaly detection \cite{zhang2024realnet, yao2022explicit,
yang2023memseg, li2023target} illustrates the performance gain by focusing on the
foreground area in the object-oriented tasks. In this paper, we also follow this
methodology to reduce the anomaly scores of uninterested background areas. This work
basically employs the binary classification strategy of the CPR algorithm
\cite{li2023target} to estimate the foreground region. However, instead of directly
predicting the foreground region on the test image, we use the union of the
foreground regions of its $k$-NN images as the foreground estimation. In this way, the
extra computation of the foreground is negligible.

\subsection{Implementation Details}

In this paper, all images are resized to $512 \times 512$, and a Wide-ResNet-50 model
\cite{zagoruyko2016wide} (pre-trained on ImageNet-1K \cite{russakovsky2015imagenet}) is
employed as $\Psi_{\mathrm{CNN}}(\cdot)$ to extract deep features. Feature maps from
layers 1, 2, and 3 are combined to form $d_f = 1024$ feature vectors, as described in
\cite{roth2022towards}. From these, $10\%$ are sampled to build the bank $\mathcal{B}$.
The parameters $\lambda_{\mathrm{P}} = 0.03$ for texture categories and
$\lambda_{\mathrm{P}} = 0.15$ for object categories are set to generate PCFs. Residual
features at $\theta = 1$ and $\theta = 2$ are concatenated and pooled to maintain the
original dimension.  Additionally, features from layer 1 of $\Psi_{\mathrm{CNN}}(\cdot)$
are applied to attain $K=64$ nearest images (see Eq.~\ref{equ:knn}), while features from
layer 2 are exploited to conduct foreground region estimation.

The Swin Transformer is trained using the Adam optimizer with a weight decay of 0.05, a
learning rate of $5 \times 10^{-5}$, and an Exponential Moving Average (EMA) decay of $0.999$. Models are trained from scratch in an unsupervised setting or with block labels. Unsupervised models are used to initialize weights with ResMixMatch \ref{alg:mixmatch},
applying a sharpening temperature of $t = 0.5$ and linearly ramping up the unlabeled loss
weight to $\lambda_\mathit{u} = 5$ over the first 400 steps of training, following the
MixMatch algorithm \cite{berthelot2019mixmatch}. The parameters of the focal loss
${\alpha_x}$, ${\alpha_u}$, ${\gamma_x}$, and ${\gamma_u}$ are set to $0.25$, $0.75$, $4$,
and $4$, respectively. To compare the final prediction map with the ground-truth label map, it is first upscaled
to the same size as the ground-truth via bilinear interpolation and then smoothed using a Gaussian kernel of $4$ \cite{roth2022towards}.

\vspace{-0.1cm}
\section{Experiments}
\label{sec:experiments}
\subsection{Experiment Setting}
In this section, extensive experiments are carried out to evaluate the proposed method,
compared with a comprehensive collection of SOTA methods including  PatchCore \cite{roth2022towards}, DRAEM
\cite{zavrtanik2021draem}, RD \cite{deng2022anomaly}, SSPCAB \cite{ristea2022self}, DMAD
\cite{liu2023diversity}, SimpleNet \cite{Liu_2023_CVPR}, DeSTSeg \cite{zhang2023destseg}, CFLOW \cite{gudovskiy2022cflow}, RD++
\cite{Tien_2023_CVPR}, M3DM \cite{wang2023multimodal}, 
 RD4AD\cite{deng2022anomaly}, UniAD\cite{you2022unified} , ReContrast\cite{guo2024recontrast}, DiAD\cite{he2024diffusion},  MambaAD\cite{he2024mambaad}, Dinomaly \cite{guo2024dinomaly}, PRN \cite{zhang2022prototypical}, BGAD \cite{yao2022explicit},
CPR \cite{li2023target}, DRA \cite{ding2022catching}, RealNet
\cite{zhang2024realnet} and AHL \cite{zhu2024anomaly}. Considering the conceptual
similarity in methodology, we also involve a SOTA method of weakly supervised
segmentation, \emph{i.e.} BoxTeacher \cite{cheng2023boxteacher}, in the comparison to
illustrate the practical advantage of WeakREST. The comparison is conducted on three benchmarks: the MVTec-AD \cite{bergmann2019mvtec} dataset, the
MVTec 3D \cite{bergmann2021mvtec} dataset and the KolektorSDD2 dataset
\cite{bovzivc2021mixed-KolektorSDD2}. 
The involved AD algorithms are measured comprehensively by four popular
threshold-inde\-endent metrics: Image-AUROC, Pixel-AUROC, PRO
\cite{bergmann2020uninformed} (Per Region Overlap) and AP  \cite{zavrtanik2021draem}
(Average Precision). The first one focuses on the precision of image-level anomaly detection
while the latter three measure the performance of anomaly localization. 

We perform all the experiments in both the unsupervised and supervised settings. In
the unsupervised scenario, only normal data can be accessed during training and synthetic
defects are artificially generated with pixel-wise labels. In the supervised AD tasks,
we randomly draw $10$ anomalous images with various defects to construct the train set and
remove them from the test set. We follow the data splitting
principle in \cite{zhang2022prototypical} and \cite{ding2022catching}. In supervised
experiments, the WeakREST model is firstly pretrained in the unsupervised condition and
then fine-tuned using the genuine defective samples. The required block-wise labels of our
methods are converted by using the method introduced in Sec.~\ref{sec:block_label} (for
pixel labels) and Sec.~\ref{sec:weaker} (for bounding-box and image-level labels).  All experiments are conducted on a single PC with one Intel i$5$-$13450$ CPU, 64G RAM and one NVIDIA RTX4090 GPU.

\subsection{Results on MVTec-AD}
\label{subsec:Results_on_MVTec_AD}

MVTec-AD \cite{bergmann2019mvtec} is the most popular AD dataset with $5,354$
high-resolution color images belonging to $5$ texture categories and $10$ object
categories. Each category contains a training set with only normal images and a test set
with various kinds of defects as well as defect-free images. We conduct the experiments on
this dataset within both unsupervised and supervised conditions.

The unsupervised AD results of the comparing algorithms on MVTec-AD
\cite{bergmann2019mvtec} are shown in Table~\ref{tab:unsupervised_results_on_MVTec}. As
shown in the table, our method achieves the highest average AP, 
and average
pixel AUROC for both texture and object categories and outperforms the unsupervised SOTA
by $0.3\%$
and $0.1\%$ respectively. Specifically, Weak\-REST ranks first on
$60\%$ ($9$ out of $15$) categories with AP metric and the ``first-ranking'' ratios for
the PRO and Pixel-AUROC are $33.1\%$ and $60\%$. As to image-level metric Image-AUROC, our
method also achieves the second-highest accuracy ($99.6\%$). As shown in Table~\ref{tab:multiclass}, WeakREST outperforms all the comparative methods at pixel-level metrics under multi-class unsupervised setting.

In addition, Table~\ref{tab:supervised_results_on_MVTec} illustrates that training with
genuine defective samples, WeakREST still ranks first for the average AD
performance evaluated by using all four metrics. In particular, our method outperforms the supervised SOTAs by $1.6\%$ on AP, $0.2\%$ on PRO and $0.1\%$ on Pixel-AUROC. The ``first-ranking'' ratios of WeakREST in the supervised scenario are $80\%$, $73.3\%$ and
$80\%$ one AP, PRO and Pixel-AUROC, respectively.
The proposed method outperforms all the SOTA methods on Image-AUROC in the supervised setting, e.g., the best result on Image-AUROC is $99.8\%$ by WeakREST. 

\begin{figure*} 
\begin{center}
  \includegraphics [width=0.8\textwidth]{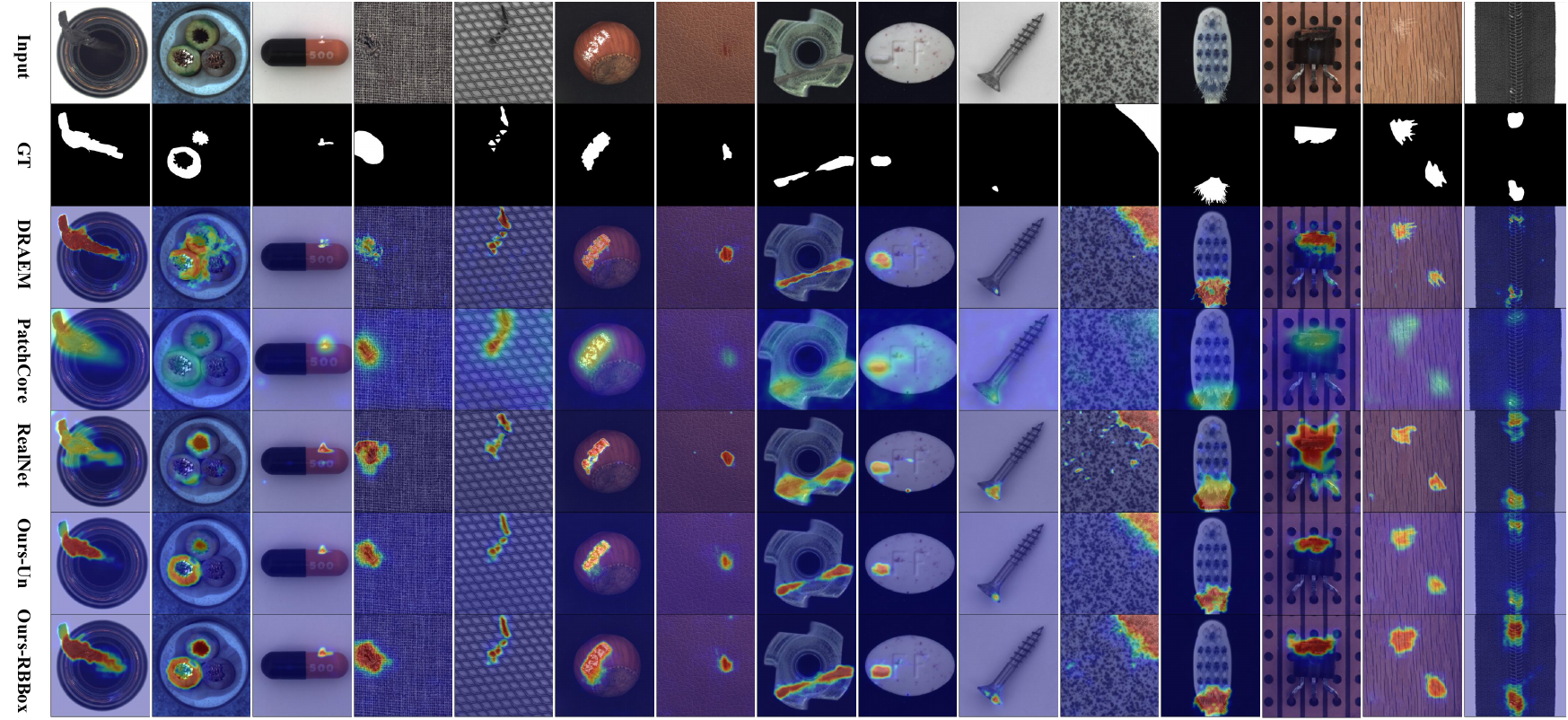}

\caption{
  Qualitative results of our WeakREST on MVTec-AD, with the two levels of supervision: Un
  (unsupervised), Weak (RBBox). Three unsupervised SOTA methods (RealNet
  \cite{zhang2024realnet}, PatchCore  \cite{roth2022towards} and DRAEM
  \cite{zavrtanik2021draem}) are also involved in the comparison.
}
\vspace*{-0.3cm}
\label{fig:qualitative}
\end{center}  
\end{figure*}

It is interesting to see that with only weak labels, WeakREST consistently outperforms existing AD algorithms with full supervision. In particular, the WeakREST learned with image tags, which requires negligible annotation cost. In contrast, the SOTA
methods need more finer labels. The proposed algorithm illustrates remarkably high capacities of exploiting the information of unlabeled regions.  
More qualitative results of the proposed method compared with other
SOTA algorithms are reported in Fig.~\ref{fig:qualitative}.

\begin{table*}
	\caption{\scriptsize
    The comparison of the Average Precision (AP), Per-Region Overlap (PRO), Pixel AUROC
    and Image AUROC metrics under unsupervised setting on the MVTec-AD dataset. The best
    accuracy in one comparison is shown in red while the second one is shown in blue.
  }
 \centering
	\resizebox{\linewidth}{!}{
\begin{tabular}{@{}lcccccccccc}
\toprule
\multirow{2}{*}{Method}& PatchCore \cite{roth2022towards}&  DRAEM \cite{zavrtanik2021draem}& NFAD \cite{yao2022explicit}& DMAD \cite{liu2023diversity}& SimpleNet \cite{Liu_2023_CVPR}& DeSTSeg \cite{zhang2023destseg}&CPR \cite{li2023target}&  RD++ \cite{Tien_2023_CVPR} &RealNet \cite{zhang2024realnet}& \multirow{2}{*}{Ours}           \\
&(CVPR2022) &(ICCV2021)  & (CVPR2023) & (CVPR2023)   &(CVPR2023)  &(CVPR2023)  & (TIP2024)  & (CVPR2023) &  (CVPR2024) &
 \\ \midrule
Carpet        &64.1/95.1/99.1&53.5/92.9/95.5&74.1/{\color{blue}{\textbf{98.2}}}/{\color{red}{\textbf{99.4}}}&63.8/95.9/99.0&44.1/92.0/97.7&72.8/$\sim$/96.1&{\color{blue}{\textbf{81.2}}}/97.6/98.9&$\sim$/97.7/99.2&62.1/96.1/98.9&{\color{red}{\textbf{81.6}}}/{\color{red}{\textbf{98.3}}}/{\color{red}{\textbf{99.4}}}\\
Grid          &30.9/93.6/98.8&{\color{blue}{\textbf{65.7}}}/{\color{blue}{\textbf{98.3}}}/{\color{red}{\textbf{99.7}}}&51.9/97.9/99.3&47.0/97.3/99.2&39.6/94.6/98.7&61.5/$\sim$/99.1&64.0/97.6/99.5&$\sim$/97.7/99.3&59.2/96.9/99.5&{\color{red}{\textbf{74.6}}}/{\color{red}{\textbf{98.7}}}/{\color{red}{\textbf{99.7}}}\\
Leather       &45.9/97.2/99.3&75.3/97.4/98.6&70.1/99.4/99.7&53.1/98.0/99.4&48.0/97.5/99.2&75.6/$\sim$/99.7&{\color{blue}{\textbf{78.5}}}/{\color{red}{\textbf{99.6}}}/{\color{red}{\textbf{99.8}}}&$\sim$/99.2/99.4&72.6/93.0/99.7&{\color{red}{\textbf{79.9}}}/{\color{blue}{\textbf{99.5}}}/{\color{red}{\textbf{99.8}}}\\
Tile          &54.9/80.2/95.7&92.3/{\color{blue}{\textbf{98.2}}}/{\color{blue}{\textbf{99.2}}}&63.0/91.8/96.7&56.5/84.3/95.8&63.5/78.3/93.9&90.0/$\sim$/98.0&{\color{blue}{\textbf{94.1}}}/98.1/{\color{blue}{\textbf{99.2}}}&$\sim$/92.4/96.6&92.2/93.7/99.1&{\color{red}{\textbf{95.4}}}/{\color{red}{\textbf{98.7}}}/{\color{red}{\textbf{99.6}}}\\
Wood          &50.0/88.3/95.0&77.7/90.3/96.4&62.9/95.6/96.9&45.5/89.3/94.8&48.8/83.9/93.9&{\color{blue}{\textbf{81.9}}}/$\sim$/97.7&80.8/{\color{red}{\textbf{97.7}}}/97.4&$\sim$/93.3/95.8&77.3/91.0/{\color{red}{\textbf{98.4}}}&{\color{red}{\textbf{84.7}}}/{\color{blue}{\textbf{97.1}}}/{\color{blue}{\textbf{98.2}}}\\\midrule
Average       &49.2/90.9/97.6&72.9/95.4/97.9&64.4/96.6/98.4&53.2/93.0/97.6&48.8/89.3/96.7&76.4/$\sim$/98.1&{\color{blue}{\textbf{79.7}}}/{\color{blue}{\textbf{98.2}}}/99.0&$\sim$/96.1/98.1&72.7/94.1/{\color{blue}{\textbf{99.1}}}&{\color{red}{\textbf{83.2}}}/{\color{red}{\textbf{98.5}}}/{\color{red}{\textbf{99.3}}}\\\midrule
Bottle        &77.7/94.7/98.5&86.5/96.8/99.1&77.9/96.6/98.9&79.6/96.4/98.8&73.0/91.5/98.0&90.3/$\sim$/99.2&{\color{blue}{\textbf{92.6}}}/{\color{red}{\textbf{98.1}}}/{\color{blue}{\textbf{99.4}}}&$\sim$/97.0/98.8&86.8/97.2/99.2&{\color{red}{\textbf{93.6}}}/{\color{blue}{\textbf{97.8}}}/{\color{red}{\textbf{99.5}}}\\
Cable         &66.3/93.2/98.4&52.4/81.0/94.7&65.7/{\color{red}{\textbf{95.9}}}/98.0&58.9/92.2/97.9&69.3/89.7/97.5&60.4/$\sim$/97.3&{\color{red}{\textbf{84.4}}}/95.2/{\color{red}{\textbf{99.3}}}&$\sim$/93.9/98.4&54.3/91.1/97.6&{\color{blue}{\textbf{84.1}}}/{\color{blue}{\textbf{95.5}}}/{\color{red}{\textbf{99.3}}}\\
Capsule       &44.7/94.8/99.0&49.4/82.7/94.3&58.7/96.0/99.2&42.2/91.6/98.1&44.7/92.8/98.9&56.3/$\sim$/99.1&{\color{blue}{\textbf{60.4}}}/{\color{blue}{\textbf{96.3}}}/{\color{red}{\textbf{99.3}}}&$\sim$/{\color{red}{\textbf{96.4}}}/98.8&59.1/90.5/{\color{red}{\textbf{99.3}}}&{\color{red}{\textbf{63.7}}}/{\color{blue}{\textbf{96.3}}}/99.2\\
Hazelnut      &53.5/95.2/98.7&{\color{red}{\textbf{92.9}}}/{\color{red}{\textbf{98.5}}}/{\color{red}{\textbf{99.7}}}&65.3/97.6/98.6&63.4/95.9/99.1&48.3/92.2/97.6&88.4/$\sim$/{\color{blue}{\textbf{99.6}}}&{\color{blue}{\textbf{88.7}}}/97.6/{\color{blue}{\textbf{99.6}}}&$\sim$/96.3/99.2&80.5/92.9/99.5&85.5/{\color{blue}{\textbf{98.2}}}/99.5\\
Metal nut     &86.9/94.0/98.3&{\color{blue}{\textbf{96.3}}}/97.0/{\color{blue}{\textbf{99.5}}}&76.6/94.9/97.7&79.0/94.2/97.1&92.6/91.3/98.7&93.5/$\sim$/98.6&93.5/{\color{blue}{\textbf{97.5}}}/99.3&$\sim$/93.0/98.1&82.1/95.1/98.1&{\color{red}{\textbf{98.3}}}/{\color{red}{\textbf{98.1}}}/{\color{red}{\textbf{99.8}}}\\
Pill          &77.9/95.0/97.8&48.5/88.4/97.6&72.6/{\color{blue}{\textbf{98.1}}}/98.0&79.7/96.9/98.5&80.1/93.9/98.5&83.1/$\sim$/98.7&{\color{red}{\textbf{91.5}}}/{\color{red}{\textbf{98.7}}}/{\color{red}{\textbf{99.5}}}&$\sim$/97.0/98.3&80.7/90.0/{\color{blue}{\textbf{99.0}}}&{\color{blue}{\textbf{84.6}}}/96.7/{\color{blue}{\textbf{99.0}}}\\
Screw         &36.1/97.1/99.5&58.2/95.0/97.6&47.4/96.3/99.2&47.9/96.5/99.3&38.8/95.2/99.2&58.7/$\sim$/98.5&{\color{red}{\textbf{71.0}}}/{\color{red}{\textbf{98.7}}}/{\color{red}{\textbf{99.7}}}&$\sim$/{\color{blue}{\textbf{98.6}}}/{\color{red}{\textbf{99.7}}}&49.2/94.0/99.4&{\color{blue}{\textbf{67.1}}}/97.3/99.5\\
Toothbrush    &38.3/89.4/98.6&44.7/85.6/98.1&38.8/92.3/98.7&71.4/91.5/99.3&51.7/88.7/98.6&75.2/$\sim$/99.3&{\color{red}{\textbf{84.1}}}/{\color{red}{\textbf{98.0}}}/{\color{red}{\textbf{99.7}}}&$\sim$/94.2/99.1&51.3/90.7/98.7&{\color{blue}{\textbf{80.8}}}/{\color{blue}{\textbf{97.2}}}/{\color{red}{\textbf{99.7}}}\\
Transistor    &66.4/92.4/96.3&50.7/70.4/90.9&56.0/82.0/94.0&58.5/85.2/94.1&69.0/93.2/96.8&64.8/$\sim$/89.1&{\color{red}{\textbf{86.7}}}/{\color{red}{\textbf{97.1}}}/{\color{red}{\textbf{98.0}}}&$\sim$/81.8/94.3&69.1/94.1/{\color{blue}{\textbf{97.6}}}&{\color{blue}{\textbf{82.5}}}/{\color{blue}{\textbf{95.3}}}/97.2\\
Zipper        &62.8/95.8/98.9&81.5/96.8/98.8&56.0/95.7/98.6&50.1/93.8/97.9&60.0/91.2/97.8&85.2/$\sim$/99.1&{\color{blue}{\textbf{88.8}}}/{\color{blue}{\textbf{98.6}}}/{\color{blue}{\textbf{99.6}}}&$\sim$/96.3/98.8&64.6/95.0/98.9&{\color{red}{\textbf{89.1}}}/{\color{red}{\textbf{98.7}}}/{\color{red}{\textbf{99.7}}}\\\midrule
Average       &61.1/94.2/98.4&66.1/89.2/97.0&61.5/94.5/98.1&63.1/93.4/98.0&62.7/92.0/98.2&75.6/$\sim$/97.9&{\color{red}{\textbf{84.2}}}/{\color{red}{\textbf{97.6}}}/{\color{red}{\textbf{99.4}}}&$\sim$/94.5/98.4&67.8/93.1/98.7&{\color{blue}{\textbf{82.9}}}/{\color{blue}{\textbf{97.1}}}/{\color{blue}{\textbf{99.2}}}\\\midrule
Total Average &57.1/93.1/98.1&68.4/91.3/97.3&62.5/95.2/98.2&59.8/93.3/97.9&58.1/91.1/97.7&75.8/$\sim$/97.9&{\color{blue}{\textbf{82.7}}}/{\color{red}{\textbf{97.8}}}/{\color{blue}{\textbf{99.2}}}&$\sim$/95.0/98.3&69.4/93.4/98.9&{\color{red}{\textbf{83.0}}}/{\color{blue}{\textbf{97.6}}}/{\color{red}{\textbf{99.3}}}\\\midrule
Image AUROC   &99.1&98.0&97.4&99.5&{\color{blue}{\textbf{99.6}}}&98.6&{\color{red}{\textbf{99.7}}}&99.4&{\color{blue}{\textbf{99.6}}}&{\color{blue}{\textbf{99.6}}}
                                              
\\\bottomrule
\end{tabular}
	}

\label{tab:unsupervised_results_on_MVTec}
\end{table*}

\begin{table*}
    \caption{\scriptsize
      The comparison of the Average Precision (AP), Per-Region Overlap (PRO), Pixel AUROC
      and Image AUROC metrics for supervised AD on the MVTec-AD dataset. The best accuracy
      in one comparison is shown in red while the
      second one is shown in blue.
    }
    \centering
       \resizebox{\textwidth}{!}{
    \begin{tabular}{lcccccccccc}
     \toprule
           \multirow{2}{*}{Method} & PRN \cite{zhang2022prototypical}& BGAD
           \cite{yao2022explicit} & CPR \cite{li2023target}& BoxTeacher
           \cite{cheng2023boxteacher} & DRA  \cite{ding2022catching} & 
           AHL \cite{zhu2024anomaly} &\multicolumn{4}{c}{ \multirow{2}{*}{Ours}}\\ 
    & (CVPR2023)  & (CVPR2023)   & (TIP2024) & (CVPR2023) & (CVPR2022) & (CVPR2024)&\\ 
    \cmidrule(lr){1-1} \cmidrule(lr){2-7} \cmidrule(lr){8-11}
    Supervision & Pixel & Pixel & Pixel & BBox & Image &Image & Block & RBBox & BBox & Image \\
    \cmidrule(lr){1-1} \cmidrule(lr){2-7} \cmidrule(lr){8-11}
Carpet        &82.0/97.0/99.0&83.2/98.9/99.6&88.1/98.9/99.6&78.3/96.4/99.2&52.3/92.2/98.2&$\sim$/$\sim$/$\sim$&{\color{blue}{\textbf{88.4}}}/{\color{red}{\textbf{99.1}}}/{\color{blue}{\textbf{99.7}}}&{\color{red}{\textbf{88.6}}}/{\color{red}{\textbf{99.1}}}/{\color{red}{\textbf{99.8}}}&87.9/{\color{red}{\textbf{99.1}}}/{\color{blue}{\textbf{99.7}}}&82.9/98.6/99.5\\
Grid          &45.7/95.9/98.4&59.2/{\color{blue}{\textbf{98.7}}}/98.4&67.3/{\color{blue}{\textbf{98.7}}}/{\color{blue}{\textbf{99.7}}}&60.0/97.9/99.4&26.8/71.5/86.0&$\sim$/$\sim$/$\sim$&{\color{red}{\textbf{76.7}}}/{\color{blue}{\textbf{98.7}}}/{\color{blue}{\textbf{99.7}}}&{\color{blue}{\textbf{75.6}}}/{\color{red}{\textbf{98.8}}}/{\color{red}{\textbf{99.8}}}&74.0/{\color{blue}{\textbf{98.7}}}/{\color{blue}{\textbf{99.7}}}&75.1/98.6/{\color{blue}{\textbf{99.7}}}\\
Leather       &69.7/99.2/99.7&75.5/99.5/99.8&78.0/99.5/99.8&56.2/97.3/98.6&5.6/84.0/93.8&$\sim$/$\sim$/$\sim$&{\color{red}{\textbf{85.7}}}/{\color{blue}{\textbf{99.6}}}/{\color{red}{\textbf{99.9}}}&{\color{blue}{\textbf{84.1}}}/{\color{blue}{\textbf{99.6}}}/{\color{red}{\textbf{99.9}}}&83.9/{\color{red}{\textbf{99.7}}}/{\color{red}{\textbf{99.9}}}&79.6/99.5/99.8\\
Tile          &96.5/98.2/99.6&94.0/97.9/99.3&97.2/99.0/99.7&91.7/96.8/98.7&57.6/81.5/92.3&$\sim$/$\sim$/$\sim$&97.4/{\color{red}{\textbf{99.2}}}/{\color{red}{\textbf{99.8}}}&{\color{red}{\textbf{97.7}}}/{\color{red}{\textbf{99.2}}}/{\color{red}{\textbf{99.8}}}&{\color{blue}{\textbf{97.6}}}/{\color{red}{\textbf{99.2}}}/{\color{red}{\textbf{99.8}}}&96.9/99.1/99.7\\
Wood          &82.6/95.9/97.8&78.7/96.8/98.0&{\color{blue}{\textbf{90.7}}}/98.4/{\color{red}{\textbf{99.5}}}&67.4/93.4/96.2&22.7/69.7/82.9&$\sim$/$\sim$/$\sim$&{\color{blue}{\textbf{90.7}}}/{\color{blue}{\textbf{98.5}}}/{\color{blue}{\textbf{99.3}}}&{\color{red}{\textbf{90.8}}}/{\color{red}{\textbf{98.6}}}/{\color{blue}{\textbf{99.3}}}&90.2/98.4/99.2&86.2/97.6/98.4\\ \cmidrule(lr){1-1} \cmidrule(lr){2-7} \cmidrule(lr){8-11}
Average       &75.3/97.2/98.9&78.1/98.4/99.2&84.3/98.9/99.6&70.7/96.4/98.4&33.0/79.8/90.6&$\sim$/$\sim$/$\sim$&{\color{red}{\textbf{87.8}}}/{\color{blue}{\textbf{99.0}}}/{\color{red}{\textbf{99.7}}}&{\color{blue}{\textbf{87.3}}}/{\color{red}{\textbf{99.1}}}/{\color{red}{\textbf{99.7}}}&86.7/{\color{blue}{\textbf{99.0}}}/{\color{red}{\textbf{99.7}}}&84.1/98.7/99.4\\ \cmidrule(lr){1-1} \cmidrule(lr){2-7} \cmidrule(lr){8-11}
Bottle        &92.3/97.0/99.4&87.1/97.1/99.3&{\color{blue}{\textbf{93.6}}}/{\color{red}{\textbf{98.5}}}/{\color{blue}{\textbf{99.6}}}&82.7/92.0/97.2&41.2/77.6/91.3&$\sim$/$\sim$/$\sim$&{\color{blue}{\textbf{93.6}}}/{\color{blue}{\textbf{98.3}}}/{\color{blue}{\textbf{99.6}}}&93.2/97.9/{\color{red}{\textbf{99.7}}}&92.8/97.7/{\color{blue}{\textbf{99.6}}}&{\color{red}{\textbf{93.8}}}/98.1/{\color{blue}{\textbf{99.6}}}\\
Cable         &78.9/{\color{blue}{\textbf{97.2}}}/98.8&81.4/{\color{red}{\textbf{97.7}}}/98.5&{\color{blue}{\textbf{88.1}}}/94.5/99.4&64.5/81.2/85.3&34.7/77.7/86.6&$\sim$/$\sim$/$\sim$&{\color{red}{\textbf{88.8}}}/96.0/99.4&87.6/96.8/{\color{red}{\textbf{99.5}}}&87.1/96.5/{\color{red}{\textbf{99.5}}}&84.5/95.5/99.3\\
Capsule       &62.2/92.5/98.5&58.3/96.8/98.8&65.8/96.7/99.4&48.1/83.1/91.3&11.7/79.1/89.3&$\sim$/$\sim$/$\sim$&{\color{blue}{\textbf{71.3}}}/97.5/{\color{red}{\textbf{99.5}}}&{\color{red}{\textbf{71.6}}}/{\color{red}{\textbf{98.2}}}/{\color{red}{\textbf{99.5}}}&68.7/{\color{blue}{\textbf{97.9}}}/99.4&66.8/97.2/99.4\\
Hazelnut      &{\color{blue}{\textbf{93.8}}}/97.4/{\color{blue}{\textbf{99.7}}}&82.4/98.6/99.4&{\color{red}{\textbf{94.4}}}/98.7/{\color{red}{\textbf{99.8}}}&77.4/95.4/99.5&22.5/86.9/89.6&$\sim$/$\sim$/$\sim$&86.9/{\color{blue}{\textbf{98.8}}}/{\color{blue}{\textbf{99.7}}}&87.6/{\color{red}{\textbf{99.0}}}/99.6&86.3/{\color{blue}{\textbf{98.8}}}/99.6&88.1/98.5/99.6\\
Metal nut     &98.0/95.8/99.7&97.3/96.8/99.6&98.6/{\color{red}{\textbf{98.4}}}/99.8&88.6/79.2/97.4&29.9/76.7/79.5&$\sim$/$\sim$/$\sim$&{\color{red}{\textbf{99.3}}}/98.2/{\color{red}{\textbf{99.9}}}&{\color{blue}{\textbf{98.9}}}/98.3/{\color{red}{\textbf{99.9}}}&98.8/{\color{red}{\textbf{98.4}}}/{\color{red}{\textbf{99.9}}}&98.6/98.3/99.8\\
Pill          &91.3/97.2/99.5&92.1/{\color{blue}{\textbf{98.7}}}/99.5&90.7/{\color{red}{\textbf{98.9}}}/99.5&75.2/85.8/96.4&21.6/77.0/84.5&$\sim$/$\sim$/$\sim$&93.4/97.8/{\color{blue}{\textbf{99.7}}}&{\color{red}{\textbf{94.8}}}/{\color{blue}{\textbf{98.7}}}/{\color{red}{\textbf{99.8}}}&{\color{blue}{\textbf{93.9}}}/97.7/{\color{blue}{\textbf{99.7}}}&88.7/97.3/99.5\\
Screw         &44.9/92.4/97.5&55.3/96.8/99.3&{\color{red}{\textbf{72.5}}}/{\color{red}{\textbf{98.9}}}/{\color{red}{\textbf{99.8}}}&35.3/56.8/79.6&5.0/30.1/54.0&$\sim$/$\sim$/$\sim$&{\color{blue}{\textbf{71.8}}}/98.1/{\color{blue}{\textbf{99.7}}}&71.5/{\color{blue}{\textbf{98.2}}}/{\color{blue}{\textbf{99.7}}}&70.9/97.9/{\color{blue}{\textbf{99.7}}}&70.0/97.7/99.6\\
Toothbrush    &78.1/95.6/99.6&71.3/96.4/99.5&84.8/{\color{red}{\textbf{98.0}}}/{\color{red}{\textbf{99.7}}}&41.0/72.5/94.6&4.5/56.1/75.5&$\sim$/$\sim$/$\sim$&84.8/{\color{red}{\textbf{98.0}}}/{\color{red}{\textbf{99.7}}}&85.4/97.5/{\color{red}{\textbf{99.7}}}&{\color{red}{\textbf{85.6}}}/97.5/{\color{red}{\textbf{99.7}}}&{\color{blue}{\textbf{85.5}}}/97.6/{\color{red}{\textbf{99.7}}}\\
Transistor    &85.6/94.8/98.4&82.3/97.1/97.9&88.1/98.0/98.4&32.1/52.8/70.8&11.0/49.0/79.1&$\sim$/$\sim$/$\sim$&{\color{red}{\textbf{94.0}}}/{\color{red}{\textbf{99.0}}}/{\color{red}{\textbf{99.6}}}&{\color{blue}{\textbf{88.5}}}/{\color{blue}{\textbf{98.7}}}/{\color{blue}{\textbf{99.2}}}&87.0/98.5/99.0&83.5/97.1/98.2\\
Zipper        &77.6/95.5/98.8&78.2/97.7/99.3&{\color{red}{\textbf{91.6}}}/{\color{blue}{\textbf{98.9}}}/{\color{red}{\textbf{99.8}}}&73.9/96.8/99.0&42.9/91.0/96.9&$\sim$/$\sim$/$\sim$&{\color{blue}{\textbf{91.5}}}/{\color{red}{\textbf{99.1}}}/{\color{red}{\textbf{99.8}}}&90.4/{\color{blue}{\textbf{98.9}}}/99.7&90.4/{\color{blue}{\textbf{98.9}}}/99.7&89.2/98.7/99.7\\ \cmidrule(lr){1-1} \cmidrule(lr){2-7} \cmidrule(lr){8-11}
Average       &80.3/95.5/99.0&78.6/97.4/99.1&86.8/97.9/99.5&61.9/79.6/91.1&22.5/70.1/82.6&$\sim$/$\sim$/$\sim$&{\color{red}{\textbf{87.5}}}/{\color{blue}{\textbf{98.1}}}/{\color{red}{\textbf{99.6}}}&{\color{blue}{\textbf{87.0}}}/{\color{red}{\textbf{98.2}}}/{\color{red}{\textbf{99.6}}}&86.1/98.0/{\color{red}{\textbf{99.6}}}&84.8/97.6/99.4\\  \cmidrule(lr){1-1} \cmidrule(lr){2-7} \cmidrule(lr){8-11}
Total Average &78.6/96.1/99.0&78.4/97.7/99.2&86.0/98.3/99.6&64.8/85.2/93.5&26.0/73.3/85.3&$\sim$/$\sim$/$\sim$&{\color{red}{\textbf{87.6}}}/{\color{blue}{\textbf{98.4}}}/{\color{red}{\textbf{99.7}}}&{\color{blue}{\textbf{87.1}}}/{\color{red}{\textbf{98.5}}}/{\color{red}{\textbf{99.7}}}&86.3/98.3/99.6&84.6/98.0/99.4\\  \cmidrule(lr){1-1} \cmidrule(lr){2-7} \cmidrule(lr){8-11}
Image AUROC   &99.4&99.3&{\color{blue}{\textbf{99.7}}}&83.4&95.9&97.0&{\color{red}{\textbf{99.8}}}&{\color{red}{\textbf{99.8}}}&{\color{red}{\textbf{99.8}}}&{\color{blue}{\textbf{99.7}}}\\
 \bottomrule 
    \end{tabular}
    }

\label{tab:supervised_results_on_MVTec}
\end{table*}

\begin{table*}[t]
    \caption{\scriptsize
    AP, PRO, Pixel-AUROC and Image-AUROC scores on MVTec-3D
   \cite{bergmann2021mvtec} with pure RGB inputs.
  }
  \centering

  \resizebox{\linewidth}{!}{
    \begin{tabular}{lcccccccccccc}
      \toprule
      \multirow{2}{*}{Method}   & PatchCore \cite{roth2022towards} &CDO \cite{cao2023collaborative}        & M3DM \cite{wang2023multimodal}& CPR\cite{li2023target} &\multirow{2}{*}{Ours} &BGAD
           \cite{yao2022explicit} & BoxTeacher
           \cite{cheng2023boxteacher} & DRA  \cite{ding2022catching} &    \multirow{2}{*}{Ours} &\multirow{2}{*}{Ours} &\multirow{2}{*}{Ours} &\multirow{2}{*}{Ours}                                   \\
           &(CVPR2022)& (TII2023) & (CVPR2023) & (TIP2024) & & (CVPR2023) & (CVPR2023) & (CVPR2022)\\ 
               \cmidrule(lr){1-1} \cmidrule(lr){2-6} \cmidrule(lr){7-13}
Supervision & Un &Un &Un &Un &Un & Pixel & BBox & Image & Block & RBBox & BBox & Image\\
 \cmidrule(lr){1-1} \cmidrule(lr){2-6} \cmidrule(lr){7-13}
    Bagel       &35.2/74.7/94.7&50.4/98.0/99.3&58.1/94.5/99.1&{\color{red}{\textbf{83.3}}}/{\color{red}{\textbf{99.5}}}/{\color{red}{\textbf{99.8}}}&{\color{blue}{\textbf{72.8}}}/{\color{blue}{\textbf{98.8}}}/{\color{blue}{\textbf{99.6}}}   &61.1/99.0/99.4&{\color{blue}{\textbf{79.4}}}/92.1/96.6&$\sim$/$\sim$/$\sim$&{\color{red}{\textbf{80.1}}}/{\color{red}{\textbf{99.4}}}/{\color{red}{\textbf{99.7}}}&69.3/98.3/99.6&67.0/98.0/99.5&76.4/{\color{blue}{\textbf{99.1}}}/{\color{red}{\textbf{99.7}}}\\
Cable Gland &27.9/96.4/99.0&42.7/{\color{blue}{\textbf{98.5}}}/99.4&40.6/97.6/99.4&{\color{red}{\textbf{61.5}}}/{\color{blue}{\textbf{98.5}}}/{\color{blue}{\textbf{99.6}}}&{\color{blue}{\textbf{53.0}}}/{\color{red}{\textbf{99.0}}}/{\color{red}{\textbf{99.7}}}&37.6/97.0/98.9&28.9/67.7/76.6&$\sim$/$\sim$/$\sim$&{\color{red}{\textbf{62.7}}}/99.2/{\color{red}{\textbf{99.8}}}&61.0/{\color{red}{\textbf{99.3}}}/{\color{blue}{\textbf{99.7}}}&{\color{blue}{\textbf{61.2}}}/{\color{red}{\textbf{99.3}}}/{\color{blue}{\textbf{99.7}}}&57.2/99.1/{\color{blue}{\textbf{99.7}}}\\
Carrot      &24.5/97.0/99.2&27.5/{\color{blue}{\textbf{97.9}}}/{\color{blue}{\textbf{99.4}}}&32.1/97.3/{\color{blue}{\textbf{99.4}}}&{\color{blue}{\textbf{37.5}}}/96.8/99.0&{\color{red}{\textbf{58.8}}}/{\color{red}{\textbf{99.2}}}/{\color{red}{\textbf{99.8}}} &47.1/98.8/99.6&53.1/93.4/96.7&$\sim$/$\sim$/$\sim$&{\color{blue}{\textbf{65.5}}}/99.2/{\color{red}{\textbf{99.8}}}&{\color{blue}{\textbf{65.5}}}/{\color{red}{\textbf{99.4}}}/{\color{red}{\textbf{99.8}}}&{\color{red}{\textbf{66.9}}}/{\color{red}{\textbf{99.4}}}/{\color{red}{\textbf{99.8}}}&64.2/99.3/{\color{red}{\textbf{99.8}}}\\
Cookie      &28.8/78.7/92.6&49.9/88.7/98.0&50.9/88.5/97.1&{\color{blue}{\textbf{59.8}}}/{\color{red}{\textbf{94.6}}}/{\color{blue}{\textbf{98.3}}}&{\color{red}{\textbf{62.9}}}/{\color{red}{\textbf{94.6}}}/{\color{red}{\textbf{98.7}}} &49.8/{\color{blue}{\textbf{95.4}}}/98.1&51.7/64.5/79.3&$\sim$/$\sim$/$\sim$&{\color{red}{\textbf{72.9}}}/94.1/98.5&63.7/94.8/98.7&{\color{blue}{\textbf{66.9}}}/{\color{red}{\textbf{95.5}}}/{\color{red}{\textbf{98.9}}}&64.9/94.2/{\color{blue}{\textbf{98.8}}}\\
Dowel       &36.5/95.5/99.1&44.3/97.5/99.6&51.3/97.6/{\color{blue}{\textbf{99.7}}}&{\color{blue}{\textbf{58.6}}}/{\color{blue}{\textbf{98.5}}}/{\color{blue}{\textbf{99.7}}}&{\color{red}{\textbf{65.5}}}/{\color{red}{\textbf{99.4}}}/{\color{red}{\textbf{99.8}}} &63.5/99.0/99.7&15.3/65.3/87.0&$\sim$/$\sim$/$\sim$&65.9/99.4/99.8&67.9/{\color{red}{\textbf{99.5}}}/{\color{red}{\textbf{99.9}}}&{\color{blue}{\textbf{68.0}}}/{\color{red}{\textbf{99.5}}}/{\color{red}{\textbf{99.9}}}&{\color{red}{\textbf{69.2}}}/{\color{red}{\textbf{99.5}}}/99.8\\
Foam        &15.8/79.6/93.6&20.5/68.1/87.6&33.0/84.5/{\color{blue}{\textbf{95.6}}}&{\color{red}{\textbf{52.7}}}/{\color{red}{\textbf{90.8}}}/{\color{red}{\textbf{97.3}}}&{\color{blue}{\textbf{42.7}}}/{\color{blue}{\textbf{86.2}}}/{\color{blue}{\textbf{95.6}}}  &25.6/71.7/90.0&39.8/77.0/85.3&$\sim$/$\sim$/$\sim$&{\color{red}{\textbf{50.4}}}/{\color{red}{\textbf{91.5}}}/{\color{red}{\textbf{97.3}}}&44.5/87.9/96.0&{\color{blue}{\textbf{45.5}}}/{\color{blue}{\textbf{88.2}}}/{\color{blue}{\textbf{96.1}}}&{\color{blue}{\textbf{45.5}}}/87.5/96.0\\
Peach       &14.8/85.1/96.0&51.2/{\color{blue}{\textbf{98.6}}}/99.6&44.3/97.0/99.4&{\color{red}{\textbf{65.0}}}/{\color{blue}{\textbf{98.6}}}/{\color{red}{\textbf{99.7}}}&{\color{red}{\textbf{65.0}}}/{\color{red}{\textbf{99.0}}}/{\color{red}{\textbf{99.7}}} &54.3/98.7/99.6&{\color{blue}{\textbf{68.1}}}/79.9/89.5&$\sim$/$\sim$/$\sim$&{\color{red}{\textbf{75.4}}}/{\color{red}{\textbf{99.5}}}/{\color{red}{\textbf{99.8}}}&61.3/98.8/{\color{blue}{\textbf{99.7}}}&58.3/98.5/99.6&63.8/{\color{blue}{\textbf{98.9}}}/{\color{blue}{\textbf{99.7}}}\\
Potato      &9.5/94.4/98.4&18.2/{\color{blue}{\textbf{95.3}}}/{\color{blue}{\textbf{99.1}}}&24.7/{\color{blue}{\textbf{95.3}}}/99.0&{\color{blue}{\textbf{28.4}}}/95.0/98.6&{\color{red}{\textbf{36.3}}}/{\color{red}{\textbf{98.4}}}/{\color{red}{\textbf{99.6}}}&30.2/98.5/{\color{blue}{\textbf{99.6}}}&24.3/85.2/93.8&$\sim$/$\sim$/$\sim$&{\color{red}{\textbf{49.0}}}/{\color{red}{\textbf{98.9}}}/{\color{red}{\textbf{99.7}}}&33.4/98.0/99.5&{\color{blue}{\textbf{43.1}}}/{\color{blue}{\textbf{98.7}}}/{\color{blue}{\textbf{99.6}}}&33.7/97.9/99.5\\
Rope        &49.8/96.3/99.4&41.1/96.8/99.4&50.8/94.9/99.3&{\color{blue}{\textbf{74.8}}}/{\color{blue}{\textbf{98.3}}}/{\color{blue}{\textbf{99.7}}}&{\color{red}{\textbf{78.9}}}/{\color{red}{\textbf{99.4}}}/{\color{red}{\textbf{99.8}}}&57.3/99.2/99.7&73.9/91.5/99.3&$\sim$/$\sim$/$\sim$&{\color{red}{\textbf{81.8}}}/{\color{red}{\textbf{99.6}}}/{\color{red}{\textbf{99.9}}}&{\color{blue}{\textbf{80.0}}}/{\color{blue}{\textbf{99.4}}}/{\color{red}{\textbf{99.9}}}&79.1/{\color{blue}{\textbf{99.4}}}/{\color{red}{\textbf{99.9}}}&77.9/{\color{blue}{\textbf{99.4}}}/99.8\\
Tire        &19.9/93.3/98.4&36.7/97.8/99.5&40.6/97.1/99.5&{\color{blue}{\textbf{55.8}}}/{\color{blue}{\textbf{98.6}}}/{\color{blue}{\textbf{99.7}}}&{\color{red}{\textbf{62.0}}}/{\color{red}{\textbf{99.2}}}/{\color{red}{\textbf{99.8}}}&28.6/96.8/99.2&50.0/64.9/82.1&$\sim$/$\sim$/$\sim$&62.6/99.1/99.8&{\color{red}{\textbf{63.8}}}/{\color{red}{\textbf{99.3}}}/{\color{red}{\textbf{99.9}}}&{\color{blue}{\textbf{62.9}}}/{\color{red}{\textbf{99.3}}}/{\color{red}{\textbf{99.9}}}&62.2/99.2/99.8\\ \cmidrule(lr){1-1} \cmidrule(lr){2-6} \cmidrule(lr){7-13}
Average     &26.3/89.1/97.0&38.2/93.7/98.1&42.6/94.4/98.7&{\color{blue}{\textbf{57.8}}}/{\color{blue}{\textbf{96.9}}}/{\color{blue}{\textbf{99.1}}}&{\color{red}{\textbf{59.8}}}/{\color{red}{\textbf{97.3}}}/{\color{red}{\textbf{99.2}}} &45.5/95.4/98.4&48.4/78.2/88.6&$\sim$/$\sim$/$\sim$&{\color{red}{\textbf{66.6}}}/{\color{red}{\textbf{98.0}}}/{\color{red}{\textbf{99.4}}}&61.0/97.5/{\color{blue}{\textbf{99.3}}}&{\color{blue}{\textbf{61.9}}}/{\color{blue}{\textbf{97.6}}}/{\color{blue}{\textbf{99.3}}}&61.5/97.4/{\color{blue}{\textbf{99.3}}}\\ \cmidrule(lr){1-1} \cmidrule(lr){2-6} \cmidrule(lr){7-13}
Image AUROC & 82.5& $\sim$ &85.0&{\color{blue}{\textbf{88.5}}}&{\color{red}{\textbf{90.2}}}&88.9&83.0&86.3&{\color{red}{\textbf{93.6}}}&89.9&{\color{blue}{\textbf{91.5}}}&90.7\\

      \bottomrule
    \end{tabular}
  }

  \label{table:3d_result}
\end{table*}

\begin{table}[tb]
\caption{\scriptsize
  Results of anomaly localization and detection performance on MVTec AD and MVTec 3D under ``Multi-class'' setting. 
  }
\centering
\resizebox{\linewidth}{!}{
\begin{tabular}{@{}lcccccccc@{}}
\toprule
Dataset& \multicolumn{4}{c}{MVTec AD \cite{bergmann2019mvtec}}&\multicolumn{4}{c}{MVTec 3D (RGB) \cite{bergmann2021mvtec}}\\
\cmidrule(lr){1-1} \cmidrule(lr){2-5} \cmidrule(lr){6-9}
Method               & AP   & PRO  & P-AUROC & I-AUROC & AP   & PRO  & P-AUROC & I-AUROC \\ \cmidrule(lr){1-1} \cmidrule(lr){2-5} \cmidrule(lr){6-9}
RD4AD\cite{deng2022anomaly} &48.6&91.1&96.1&94.6&29.8&93.5&98.4&77.9\\
SimpleNet\cite{Liu_2023_CVPR}&45.9&86.5&96.8&95.3&18.3&77.6&93.5&72.5\\
DeSTSeg \cite{zhang2023destseg}&54.3&64.8&93.1&89.2&38.1&46.4&95.1&79.6\\
UniAD\cite{you2022unified} &43.4&90.7&96.8&96.5&21.2&88.1&96.5&78.9\\
DiAD\cite{he2024diffusion}&52.6&90.7&96.8&97.2&25.3&87.8&96.4&84.6\\
MambaAD\cite{he2024mambaad}&56.3&93.1&{\color{blue}{\textbf{97.7}}}&{\color{blue}{\textbf{98.6}}}&37.5&93.6&98.6&{\color{blue}{\textbf{86.2}}}\\
Dinomaly \cite{guo2024dinomaly}& {\color{blue}{\textbf{68.7}}}&{\color{blue}{\textbf{94.7}}}&{\color{red}{\textbf{98.3}}}&{\color{red}{\textbf{99.6}}} & {\color{blue}{\textbf{55.0}}}&{\color{red}{\textbf{96.5}}}&{\color{red}{\textbf{99.2}}}&{\color{red}{\textbf{90.6}}}\\
CPR  \cite{li2023target} &  63.3& 93.1&97.2& 95.7 & 37.6& 95.1&98.4& 80.9 \\
Ours    & {\color{red}{\textbf{77.1}}}& {\color{red}{\textbf{95.4}}}& {\color{red}{\textbf{98.3}}}&98.5 & {\color{blue}{\textbf{48.8}}}& {\color{blue}{\textbf{95.4}}}& {\color{blue}{\textbf{98.6}}}& 83.8\\ 
\bottomrule
\end{tabular}
}

\label{tab:multiclass}
\end{table}

\subsection{Results on MVTec 3D}
\label{subsec:Results_on_MVTec_3D}
As a more challenging alternative to MVTec-AD, MVTec 3D  \cite{bergmann2021mvtec}  contains over $4000$ high-resolution color images and 3D point cloud data of ten industrial products.
Each product includes normal images in the train set and the corresponding test set
consists of both defective and defect-free images. 

We evaluate our algorithm on the MVTec 3D dataset with those SOTA methods also reporting their
results on this dataset. Table~\ref{table:3d_result} shows that WeakREST achieves
better performances to the unsupervised and supervised SOTA. In the unsupervised condition, the proposed method
surpasses SOTA methods by large margins: $2.0\%$, $0.4\%$, $0.1\%$ and $1.7\%$ on AP,
PRO, Pixel-AUROC and Image-AUROC, respectively. WeakREST also demonstrates better performance under multi-class setting, shown in Table~\ref{tab:multiclass}. Similar to the situation of MVTec-AD, the
weakly-supervised WeakREST models also obtains higher average performances than the
fully-supervised SOTA algorithms.

\subsection{Results on KolektorSDD2}
\label{subsec:Results_on_KolektorSDD2}
KolektorSDD2  \cite{bovzivc2021mixed-KolektorSDD2} dataset is designed for surface defect
detection and includes various types of defects, such as scratches, minor spots, and
surface imperfections. It comprises a training set with $246$ positive (defective) and
$2,085$ negative (defect-free) images, as well as a test set with $110$ positive and $894$
negative images. We compare the performances of WeakREST with the SOTA results available
in the literature. 

As shown in Table~\ref{tab:supervised_KSDD2}, the unsupervised WeakREST beats SOTA methods
with a clear superiority ($12.3\%$, $3.4\%$, $2.0\%$ and $0.8\%$ for AP, PRO, Pixel-AUROC and
Image-AUROC, respectively). Under the supervised condition, our method also achieves better
results and the WeakREST model supervised by image labels can already outperform existing
methods with pixel-wise annotations.

\begin{table}
\caption{\scriptsize
  Results of anomaly localization performance on KolektorSDD2. 
  The upper sub-table shows the results obtained in the
  unsupervised condition and the lower part reports those with genuine defective samples.}
\centering
\resizebox{0.8\linewidth}{!}{
\begin{tabular}{@{}lccccc@{}}
\toprule
Method& Supervision                & AP   & PRO  & P-AUROC & I-AUROC \\ \midrule
PatchCore \cite{roth2022towards}   &Un    & {\color{blue}{\textbf{64.1}}} & 88.8 & 97.1       & 94.6  \\
DRAEM \cite{zavrtanik2021draem}      &Un         & 39.1 & 67.9 & 85.6      &81.1   \\
SSPCAB \cite{ristea2022self}         &Un          & 44.5 & 66.1 & 86.2   & 83.4      \\
CFLOW \cite{gudovskiy2022cflow}        &Un           & 46.0   & 93.8 & 97.4  & 95.2       \\
RD \cite{deng2022anomaly}                  &Un     & 43.5 & {\color{blue}{\textbf{94.7}}} & {\color{blue}{\textbf{97.6}}}   & {\color{blue}{\textbf{96.0}}}      \\
Ours &   Un& {\color{red}{\textbf{76.4}}} & {\color{red}{\textbf{98.1}}} & {\color{red}{\textbf{99.6}}} &  {\color{red}{\textbf{96.8}}}       \\\midrule
PRN \cite{zhang2022prototypical} & Pixel     &72.5& 94.9 & {\color{blue}{\textbf{97.6}}}  &96.4       \\
Box2Mask \cite{chibane2022box2mask} &BBox & 35.3&74.8&79.2 &86.1\\
BoxTeacher \cite{cheng2023boxteacher}& BBox & 23.2&79.3&90.9&74.9\\
Ours &Block     & {\color{red}{\textbf{77.7}}} & {\color{red}{\textbf{99.0}}} & {\color{red}{\textbf{99.7}}} & {\color{red}{\textbf{97.9}}}         \\
Ours &RBBox & 76.9 & {\color{blue}{\textbf{98.9}}} & {\color{red}{\textbf{99.7}}}      &97.5   \\
Ours &BBox &76.4& 98.8& {\color{red}{\textbf{99.7}}}  &97.6       \\
Ours &Image & {\color{blue}{\textbf{77.0}}} & 98.7 & {\color{red}{\textbf{99.7}}}  & {\color{blue}{\textbf{97.7}}}        \\
\bottomrule
\end{tabular}
}

\label{tab:supervised_KSDD2}
\end{table}

\begin{table}
  \caption{\scriptsize
    The impact of the block-label thresholds (defined in Eq.~\ref{equ:block_label}). The
    test is performed on MVTec-AD using AP, PRO, Pixel-AUROC, and Image-AUROC metrics in
    both unsupervised and supervised scenarios.
  }
\centering
  \resizebox{0.8\linewidth}{!}{
  \begin{tabular}{@{}cccc@{}}
  \toprule
  $\epsilon^{+}$ & $\epsilon^{-}$ & Unsupervised                                                                                                                & Weak-sup (RBBox)                                                                                                            \\ \midrule
  0.25                & 0.00                   & 82.8/97.5/99.2/99.5                                                                                                         & \textcolor{red}{\textbf{87.1}}/98.4/99.6/\textcolor{red}{\textbf{99.8}}                                                     \\
  0.50                 & 0.10                 & \textcolor{red}{\textbf{83.0}}/\textbf{\textcolor{red}{97.6}}/\textbf{\textcolor{red}{99.3}}/\textcolor{red}{\textbf{99.6}} & \textcolor{red}{\textbf{87.1}}/\textcolor{red}{\textbf{98.5}}/\textbf{\textcolor{red}{99.7}}/\textcolor{red}{\textbf{99.8}} \\
  0.75                & 0.20                 & 82.4/\textcolor{red}{\textbf{97.6}}/\textcolor{red}{\textbf{99.3}}/\textcolor{red}{\textbf{99.6}}                           & 86.8/\textbf{\textcolor{red}{98.5}}/99.6/\textbf{\textcolor{red}{99.8}}                              \\\bottomrule
  \end{tabular}
  }

\label{tab:block}
\end{table}

\begin{table}
  \caption{\scriptsize
    Bounding-box label perturbation analysis. The first column denotes the scales of the
    perturbation The test is conducted on MVTec AD with AP, PRO, Pixel-AUROC, and
    Image-AUROC metrics.}
\centering
  \resizebox{0.8\linewidth}{!}{
  \begin{tabular}{@{}ccc@{}}
  \toprule
  Perturb. (pixel)     & RBBox & BBox \\ \midrule
  0&87.1/98.5/99.7/99.8	&86.3/98.3/99.6/99.8	\\
  $-3\sim+3$&86.7/98.5/99.6/99.8&	86.2/98.3/99.6/99.7	\\
  $-5\sim+5$&86.8/98.5/99.6/99.8	&85.3/98.2/99.6/99.8	\\
  $-7\sim+7$&86.0/98.4/99.6/99.8&	85.9/98.2/99.6/99.8	\\\bottomrule
  \end{tabular}
  }

\label{tab:perturbation}
\end{table}

\begin{table}
  \caption{\scriptsize
    Evaluation on backbone selection on MVTec AD across AP, PRO, Pixel-AUROC, and Image-AUROC
    metrics in both unsupervised and weakly-supervised scenarios.
  }
  \centering
  \resizebox{0.8\linewidth}{!}{
  \begin{tabular}{@{}lcc@{}}
  \toprule
  Backbone                & Unsupervised  & Weak-sup (RBBox) \\ \midrule
  Swin \cite{liu2021swin} &  {\color{red}{\textbf{83.0}}}/{\color{red}{\textbf{97.6}}}/{\color{red}{\textbf{99.3}}}/{\color{red}{\textbf{99.6}}} &  {\color{red}{\textbf{87.1}}}/{\color{red}{\textbf{98.5}}}/{\color{red}{\textbf{99.7}}}/{\color{red}{\textbf{99.8}}}
   \\
   ViT \cite{dosovitskiyimage}  & 75.7/94.0/98.5/99.3 & 80.5/96.8/99.0/99.3\\
  DeSTSeg \cite{zhang2023destseg} & 79.9/94.8/98.2/99.5
   & 74.4/89.4/97.5/99.0
  \\\bottomrule
  \end{tabular}
  }

\label{tab:Backbone}
\end{table}

\begin{figure}[!tb]
  \begin{center}
  \includegraphics [scale=0.6]{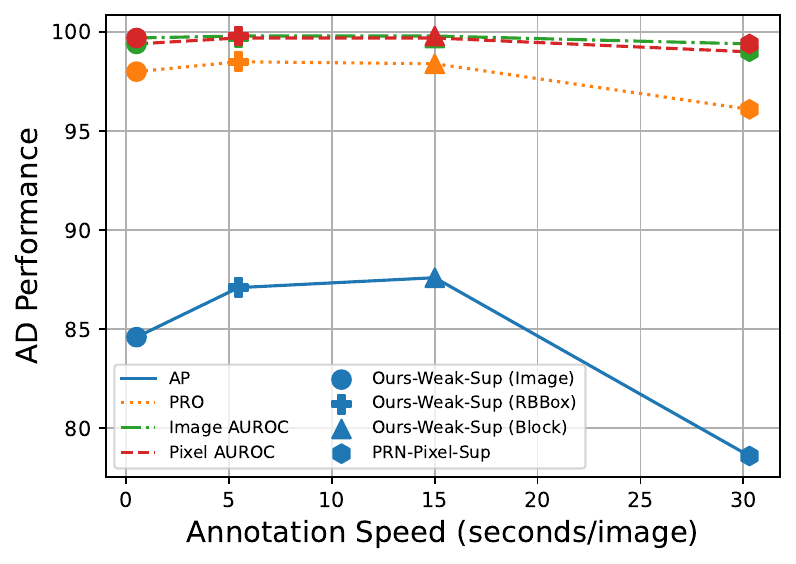}
  \caption{\scriptsize
      The per-image annotation costs (x-axis) of the three levels of anomaly labels are
      shown as the circle (image label) plus (bounding-box label), triangle (block-wise
      label) and pentagon (pixel-wise label) shapes. The y-axis stands for the AD
      performances with the four metrics, shown as red-dashed (Pixel-AUROC), green-dashed
      (Image-AUROC), orange-dot (PRO) and blue-solid (AP) lines.      
  }
  \vspace*{-0.1cm}
\label{fig:label_time}
\end{center}  
\end{figure}

\begin{table*}
\caption{\scriptsize
    Ablation study results of the WeakREST algorithm on MVTec AD. 
}
\centering
  \resizebox{0.8\textwidth}{!}{
  \begin{tabular}{cccccccccccccc}
  \toprule
  \multirow{2}{*}{Setting}&\multicolumn{8}{c}{Module} & \multicolumn{5}{c}{Performance} \\ 
  \cmidrule(lr){2-9}\cmidrule(lr){10-14}
  &Swin   &PCF &PCA       & Filter & Fore   & ResMix   &Masks & Jitter & AP & PRO & P-AUROC   & I-AUROC & Latency (ms)                                                       \\
   \cmidrule(lr){1-1}\cmidrule(lr){2-9}\cmidrule(lr){10-14}
  \multirow{6}{*}{Un}&   &   &   &   &   &   &&  &66.2&95.0&97.6&96.7&65.4\\
  &\checkmark&   &   &   &   &   &   &  &79.5&96.8&98.6&99.3&79.1\\
  &\checkmark&\checkmark &   &   &   &   &   &   &82.8&{\color{red}{\textbf{97.8}}}&{\color{red}{\textbf{99.4}}}&99.5&79.5\\
  &\checkmark&\checkmark & \checkmark   &   &   &&&&82.1&97.7&99.3&99.4
     &56.1\\
  &\checkmark&\checkmark   &  \checkmark & \checkmark   &   &   &   &&82.6&97.7&99.3&99.3
  &{\color{red}{\textbf{39.4}}}\\
  &\checkmark&\checkmark & \checkmark   & \checkmark  & \checkmark  &   &   &   &{\color{red}{\textbf{83.0}}}&97.6&99.3&{\color{red}{\textbf{99.6}}} &39.7
  \\ \cmidrule(lr){1-1}\cmidrule(lr){2-9}\cmidrule(lr){10-14}
   
  \multirow{4}{*}{RBBox}&\checkmark&\checkmark & \checkmark   & \checkmark  & \checkmark  &   &   &   &83.1&97.6&99.3&99.7& 39.7 \\
  &\checkmark&\checkmark & \checkmark   & \checkmark  & \checkmark  & \checkmark  &   &   &85.8&98.3&99.6&99.7
  & 39.7 \\
  &\checkmark&\checkmark & \checkmark   & \checkmark  & \checkmark  & \checkmark  &   \checkmark & &86.8&98.4&99.6&99.7
  & 39.7 \\
  &\checkmark&\checkmark & \checkmark   & \checkmark  & \checkmark  & \checkmark  &   \checkmark &\checkmark & {\color{red}{\textbf{87.1}}}& {\color{red}{\textbf{98.5}}}& {\color{red}{\textbf{99.7}}}& {\color{red}{\textbf{99.8}}}
  & 39.7 \\
   \bottomrule
  \end{tabular}
  }

\label{tab:ablation}
\end{table*}

\subsection{Analysis on weak labels}
\label{subsec:Analysis_on_weak_labels}

Recall that the main contribution of this work is to reduce the labeling cost in AD, we report the annotation time-consumption of the proposed two weak labels compared with pixel-level annotations. To clock the labeling time, the pixel labels, block labels, bounding boxes and the image
tags of anomalies on a subset of MVTec-AD ($10$ defective images for each sub-category) are
all manually annotated. Four master students majoring in computer vision completed the
labeling task using the labeling tool proposed in this work. The average annotation times of four kinds of labels are illustrated in
Fig. \ref{fig:label_time}, along with the corresponding AD performances (Image-AUROC, Pixel-AUROC, PRO and AP). As shown in Fig. \ref{fig:label_time}, one requires only around $0.5$ seconds
to label a defective image. Besides, it takes around $5$ seconds and $17$ seconds to label bounding boxes and block-wise labels on an image, respectively. In contrast, the
SOTA method \cite{zhang2022prototypical} based on pixel labels needs more than $32$ seconds for labeling one image, while yielding consistently lower accuracy. 

Recall that our block-labels are all converted from the pixel-labels based on two
pre-defined parameters $\epsilon^{+}$ and $\epsilon^{-}$ (Eq.~\ref{equ:block_label}), we
carry out an experiment to verify the model robustness on the fluctuation of these
parameters. As the results shown in Table~\ref{tab:block}, the AD accuracies of WeakREST
are generally stable when $\epsilon^{+}$ and $\epsilon^{-}$ changes significantly.

It is inevitable to introduce noise during bounding-box annotation by hands. In this regard, we test the proposed algorithm with perturbed bounding boxes and report the results in
Table~\ref{tab:perturbation}. It can be seen that even contaminated by the
noise up to $\pm7$ pixels, which is around $15\%$ of the average size of the bounding-boxes, the performance drop is negligible: around $1\%$ on AP while less than $0.2\%$ for other metrics.

\subsection{Ablation study}
\label{subsec:Ablation_study}

In this section, the contributing modules of Weak\-REST are evaluated in ablative view. The modules include: Swin Transformer
introduced in Sec.~\ref{subsubsec:swin} ({\bf Swin}); Position Constrained Feature (see Sec.~\ref{subsubsec:pcf} ({\bf PCF}); the PCA for faster matching 
(see Sec.~\ref{subsubsec:far}, {\bf PCA}); the filtering process for reference images (see Sec.~\ref{subsubsec:similar}, {\bf Filter}); the foreground estimation
proposed in Sec.~\ref{subsec:fore} ({\bf Fore}); the ResMixMatch algorithm introduced in
Sec.~\ref{subsec:mixmatch} ({\bf ResMix}); the randomly masking ({\bf Mask}) and residual
jittering ({\bf Jitter}) augmentation strategy defined in Sec.~\ref{subsubsec:mask}. From
Table~\ref{tab:ablation} we can see that most modules can improve the performance steadily except the
``PCA'' module which slightly reduce the AD performances. However, the accelerating module
increase the running speed by around $28\%$ (from $79.5$ ms to $56.1$ ms). The two
accelerating module ``PCA'' and ``Filter'' can jointly {\bf double} the algorithm speed
while keeping the accuracy nearly unchanged. 

In addition, the impact of the backbone selection over Swin Transformer
\cite{liu2021swin}, ViT \cite{dosovitskiyimage} and the segmentation network employed in
\cite{zhang2023destseg}) is illustrated in Table~\ref{tab:Backbone}. One can see that the
combination of Swin Transformer achieves the best scores while the ViT model performs
worst in the unsupervised condition, probably due to the model overfitting to the
synthetic defects. However, when genuine defective samples become available in training,
ViT surpasses the segmentation network of DeSTSeg due to its capacity for feature extraction.

\section{Conclusion}
\vspace{-0.2cm}
\label{sec:conclusion}
In this paper, we tackled the anomaly detection (AD) problem via a novel block-wise
classification, which requires much less annotation effort than the pixel-wise
segmentation. To achieve this, we designed a novel residual feature to represent various
anomaly status of the image blocks. A Swin Transformer model, learned through a novel
training strategy, classifies each block as defective or defect-free based on their
residual features. Furthermore, when using weaker labels such as bounding boxes and image
tags to roughly define defective regions, our ResMixMatch scheme effectively
exploits information from unlabeled regions, achieving AD performance close to that
obtained with strong supervision. The proposed WeakREST algorithm sets SOTA performance in the literature while requiring non-expert annotations. This work paves a way to reduce annotation costs for AD while maintaining
high accuracy. According to our experiments, the weakly-supervised setting is proven to be more practical alternative to the supervised setting that limits the number of training images. In future, we anticipate the development of even better
weakly-supervised AD algorithms by exploiting more useful information from unlabeled image
regions.

\vspace{-0.1cm}
\section{Acknowledgment}
This work was supported by National Natural Science Foundation of China (Grant No. 62372150).

\bibliography{hanxi_bib_semiRest}
\bibliographystyle{IEEEtran}

\end{document}